\documentclass[12pt,journal,onecolumn]{IEEEtran}
\IEEEoverridecommandlockouts
\usepackage{nopageno}
\usepackage[margin=2.5cm]{geometry}

\ifCLASSOPTIONcompsoc
  \usepackage[nocompress]{cite}
\else
  \usepackage{cite}
\fi
\usepackage{amsmath,amssymb,amsfonts}
\usepackage{algorithmic}
\usepackage{graphicx}
\usepackage{textcomp}
\usepackage{xcolor}
\usepackage{booktabs}
\usepackage{slashbox}
\def\BibTeX{{\rm B\kern-.05em{\sc i\kern-.025em b}\kern-.08em
    T\kern-.1667em\lower.7ex\hbox{E}\kern-.125emX}}

\usepackage[commentmarkup=uwave]{changes}
\definechangesauthor[name={Rafael Molina}, color=blue]{rms}

\definecolor{ao}{rgb}{0.0, 0.5, 0.0}
\definechangesauthor[name={Gabriel Maciá}, color=ao]{gmf}

\definecolor{pinegreen}{rgb}{0.0, 0.47, 0.44}
\definechangesauthor[name={Luz Garcia}, color=pinegreen]{lgm}

\definecolor{correccionfpb}{RGB}{255,40,120}
\definechangesauthor[name={Fernando Perez}, color=correccionfpb]{fpb}

\usepackage{bm}
\usepackage{color}
\usepackage{amssymb}

\usepackage{algorithmic}
\usepackage{algorithm}

\usepackage{multirow}

\def\p{{\mathrm{p}}}

\def\T{{\mathrm{T}}}

\def\bzero{{\mathbf 0}}

\def\bsigma{{\boldsymbol{\sigma}}}

\def\bp{{\mathbf p}}

\def\bx{{\mathbf x}}

\def\bz{{\mathbf z}}

\def\bC{{\mathbf C}}

\def\bE{{\mathbf E}}

\def\bI{{\mathbf I}}

\def\bL{{\mathbf L}}
\def\bM{{\mathbf M}}

\def\bR{{\mathbf R}}
\def\bS{{\mathbf S}}
\def\bT{{\mathbf T}}
\def\bU{{\mathbf U}}
\def\bV{{\mathbf V}}
\def\bW{{\mathbf W}}
\def\bX{{\mathbf X}}

\usepackage[utf8]{inputenc}
\usepackage[english]{babel}

\usepackage{amsthm}
\newtheorem{theorem}{Theorem}

\begin{document}

\title{\rule{\textwidth}{2pt}\\Leveraging a Probabilistic PCA model to Understand the Multivariate Statistical Network Monitoring Framework for Network Security Anomaly Detection\\\rule{\textwidth}{2pt}

}



\author{Fernando Pérez-Bueno,
        Luz García,
        Gabriel Maciá-Fernández,
        Rafael Molina
\IEEEcompsocitemizethanks{\IEEEcompsocthanksitem Fernando Pérez-Bueno (corresponding author) and Rafael Molina are with Dpto. de Ciencias de la Computación e I. A., Universidad de Granada, Spain.
E-mail: fpb@ugr.es; rms@decsai.ugr.es
\IEEEcompsocthanksitem Luz García and Gabriel Maciá-Fernández are with Dpto. de Teoria de la Señal, Telematica y Comunicaciones, Universidad de Granada, Spain.
E-mail: luzgm@ugr.es; gmacia@ugr.es
\IEEEcompsocthanksitem Preprint submitted to IEEE Transactions on Networking
}
}

\maketitle
\noindent\rule{\linewidth}{1pt}\\\par
\begin{abstract}
Network anomaly detection is a very relevant research area nowadays, especially due to its multiple applications in the field of network security. The boost of new models based on variational autoencoders and generative adversarial networks has motivated a reevaluation of traditional techniques for anomaly detection. It is, however, essential to be able to understand these new models from the perspective of the experience attained from years of evaluating network security data for anomaly detection. 
In this paper, we revisit anomaly detection techniques based on PCA from a probabilistic generative model point of view, and contribute a mathematical model that relates them. Specifically, we start with the probabilistic PCA model and explain its connection to the Multivariate Statistical Network Monitoring (MSNM) framework. MSNM was recently successfully proposed as a means of incorporating industrial process anomaly detection experience into the field of networking.
We have evaluated the mathematical model using two different datasets. The first, a synthetic dataset created to better understand the analysis proposed, and the second, UGR'16, is a specifically designed real-traffic dataset for network security anomaly detection. We have drawn conclusions that we consider to be useful when applying generative models to network security detection.
\end{abstract}


\section{Introduction}
\label{sec:intro}
Network security constitutes mainly a research and development focus nowadays, with a forecasted market of \$170.4 billion in 2022 according to Gartner \cite{Gartner2018:Online}, and a constant flow of worrying news concerning security incidents, like data breaches, denial of service, data exfiltration, privacy issues, advanced persistent threats, or even government cyberwar issues \cite{nytimes2021:online}.  

Many security technologies have been developed in recent years to deal with these relevant problems. The in-depth security design paradigm advocates the use of different layers of security protection to deal with such problems. Among these layers, the detection of network security incidents is a crucial part of developing effective response measures when attacks occur. 

There are two main approaches to the detection of network security incidents. One, \textit{signature detection} works with rules defined by experts that identify known attacks. This approach works efficiently on known attacks but lacks the flexibility to detect unseen attack patterns or multi-stage attacks. An alternative approach, known as \textit{anomaly detection}, based on building a normality model and detecting deviations from it, has gained popularity. However, the main limitation of anomaly detection technologies is related to the appearance of false positives or negatives, due to the lack of accuracy in the normality models obtained. 

In this context, many different approaches to anomaly detection have been proposed in the literature \cite{Nassif_survey}. Recent research in the deep learning area is generating high expectations with regard to the promising results that could be obtained in the field of network anomaly detection \cite{kwon2019survey}\cite{chalapathy2019deep}\cite{ruff2021unifying}. Preliminary results on the use of generative models, like variational autoencoders or generative adversarial networks, show high performance. There is, however, a concern that the internals of these models are not completely understood and that they might only show good results for specific datasets. In summary, there is a need to comprehensively understand the evolution from widely used models to these new deep generative models. 

A family of models formed by those based on PCA are currently being successfully applied in the network anomaly detection problem. The first proposal to use PCA was that of Lakhina et al. \cite{lakhina2004diagnosing}\cite{lakhina2004characterization} in 2004. A training data matrix was built with different features extracted from network flows. The latent PCA model was then extracted from it and subsequently used to evaluate new network samples to decide on their abnormality.  

Later in 2016, Camacho \textit{et al.} \cite{macia2016hierarchical} proposed the use of a framework based on PCA called MSNM (\textit{Multivariate Statistical Network Monitoring}). MSNM essentially adapts a methodology known as MSPC (Multivariate Statistical Process Control), which has been extensively and successfully used in the field of anomaly detection in industrial process control to deal with the specificities of the network anomaly detection problem. In brief, the framework suggests the construction of a PCA model\footnote{The use of linear PCA techniques is justified by the advantages in the diagnosing of anomalies in posterior phases of an incident lifecycle.} for normal traffic patterns, and the use of two statistics, termed $D$ and $Q$, which are thresholded in order to determine if an anomaly occurs. While the statistic $Q$ is similar to the one used by Lakhina, the use of $D$, a Hotelling's T2 statistic \cite{Hotelling}, is a novelty that also captures possible deviations from the latent model, rather than only in the observation model ($Q$ statistic).

Despite the effectiveness of MSNM in anomaly detection (see the comparative analysis in~\cite{Camacho2019}), two problems appear in the application of this model. First, there is a lack of understanding on how $D$ and $Q$ statistics behave in different scenarios and why. This implies that obtaining good results in certain scenarios could lead to false conclusions being drawn. Second, there is a need to relate these PCA based anomaly detection models to more novel approaches like generative models, e.g., VAEs, which are currently being used profusely in the anomaly detection area. Understanding this relationship allows well-known lessons from PCAs to be applied to generative models.

A better way to understand these models from a generative perspective was provided by Tipping and Bishop \cite{tipping1999probabilistic}, in which the PCA model is derived from a probabilistic PCA (PPCA) model when the variance of the latent distribution becomes zero. In addition, PPCA is at the basis of the definition of VAEs \cite{kingma2014}.


Starting from the PPCA model (a generative model that explains PCA), the focus of this paper is to derive an analytical model that elucidates why the use of the two MSNM statistics, $D$ and $Q$, leads to effective anomaly detection and how this is understood in the framework of generative models. Note that the aim of this paper is not to contribute with a novel approach derived from MSNM, but rather to relate this model with generative models (analytically and empirically). 

As it will be shown, while the $Q$ statistic measures the quality of the reconstruction model, the $D$ statistic will represent a regularization term in the generative model. As previously discussed, these conclusions should help to understand how to soundly employ generative models, like VAEs and GANs, for network anomaly detection. As an example, some of the current proposals for anomaly detection using VAEs, \textit{e.g.}, \cite{Xu2018VAE}, use only the reconstruction model, thus discarding the contributions of the regularization term that could be relevant, as we will show in what follows.

In summary, the contribution of this paper is threefold: \textit{i)} A PPCA model is leveraged to understand the MSNM framework from a generative point of view. We then develop a mathematical framework to explain, from a probabilistic point of view, the meaning of the $Q$ and $D$ statistics.\textit{ ii)} Using the generative model, we show how some limitations of the MSNM model are circumvented. Specifically, the authors of \cite{camacho2017traffic} propose the use of a combined weighted statistic for both $Q$ and $D$ (called the t-Score). Then, we obtain a probabilistic interpretation for this weighted combination with PPCA-MSNM. \textit{iii)} We test the generative model on both a synthetic dataset and a real network traffic dataset to show how detection results stay coherent with both models. 

The rest of this paper is structured as follows. In Section~\ref{sec:related}, we present related work and explain the contributions of this paper in more detail. Then, we provide the basics for MSNM and PPCA in Sections~\ref{sec:MSNM} and~\ref{sec:PPCA}, respectively. The mathematical model that connects MSNM and PPCA is proposed in Section~\ref{sec:PPCA_MSPC}. This proposal is validated with the experiments shown in Section~\ref{sec:experiments}. Finally, conclusions are drawn in Section~\ref{sec:conclusions}. 

\section{Related work}
\label{sec:related}


Many statistical strategies for anomaly detection on the basis of Lakhina's approximation \cite{lakhina2004diagnosing} \cite{lakhina2004characterization} have been profusely proposed. The benefits of PCA's unsupervised nature have motivated the appearance of a wide range of work, like the PCA-based traffic matrix estimation of \cite{Soule2005PCA}, the network anomography proposed by \cite{Zhang2005PCA}, or the combination of distributed tracking and in-network PCA-based anomaly detection of \cite{Huang2006PCA} among many others. Limitations of these models, like the high sensitivity to calibration settings, ineffective detection of large anomalies or difficulties to capture temporal correlations have been reported \cite{Ringberg2007LimitationsPCA}. In addition, the proposals to solve these limitations also use different frameworks. Robust PCA \cite{Rubistein2009RobustPCA}, and its variation \cite{Wang2012PCA}~\cite{Pascoal2012RobustPCA}, or the Karhunen-Loève expansion used by \cite{Brauckhoff2007LimitationsPCA}, are examples of the achieved progress.

In 2016, Camacho \textit{et al.} proposed the use of the Multivariate Statistical Network Monitoring (MSNM) framework \cite{macia2016hierarchical} \cite{camacho2017traffic} as an improvement to previous PCA proposals. In essence, MSNM is an adaptation from a sibling framework traditionally used in the field of industrial process control, known as MSPC (Multivariate Statistical Process Control) \cite{K&MG96} \cite{ferrer_latent:2014} \cite{G&F&W11}. In order to face the particularities of the networking field, MSNM adapted the MSPC methodology to introduce new data pre-processing strategies and processing steps, like the deparsing of network traces \cite{Camacho2019}. In addition, MSNM research has focused on the evaluation of its implementation in real networks, the optimization of its parameters with semi-supervised models, enabling big-data processing, its application to hierarchical architectures for issuing privacy and traffic reduction \cite{macia2016hierarchical}, enhancing visualization of network anomalies or supporting authentication systems \cite{Soufiane2021}. 

The use of deep generative models in the field of anomaly detection, is currently a hot research topic due to good performance achieved by the use of deep learning techniques. Many authors have followed this approach using variational autoencoders (VAEs) as a natural evolution of PCA in the frame of reconstruction approaches to detect anomalies. Despite the fact that image and video processing research was the initial promoter of these models~\cite{ruff2021unifying}, they  are also being extensively evaluated in the field of network anomaly detection. A recent survey of applications and techniques can be found in \cite{kwon2019survey}, yet, it is surprising to see that in many of these studies, like \cite{Xu2018VAE}, where they consider the time gradient effect, or in the conditional VAE implemented in \cite{LopezMartin2017CVAE}, or in others like \cite{An2015VAE} or \cite{Zavrak2020VAE}, anomaly scores are evaluated by only considering the reconstruction error provided by the VAE. That is, the regularization term present in the VAE marginal likelihood is not taken into account. Very few proposals, like \cite{Solch2016VAE}, use both the regularization and the error reconstruction terms of the marginal likelihood to evaluate anomaly scores.

As will be shown in the rest of this paper, not using the regularization term in VAEs is similar to the well-known limitation in Lakhina's PCA approaches that do not consider the latent variable space deviations for anomaly detection. Thus, although evaluating the impact of the use of both terms (regularization and reconstruction) in VAEs is out of the scope of this paper, the goal here is to show the connection between these terms in a PPCA generative model (precursor of VAE) and MSNM detection statistics.

\section{MSNM for anomaly detection}
\label{sec:MSNM}
Multivariate Statistical Network Monitoring (MSNM) \cite{macia2016hierarchical} transfers the theory of Statistical Processes Control, which has been used for a long time in industrial applications, to network traffic analysis. Its goal is to jointly analyze several interrelated variables to differentiate common from special causes of variation called anomalies. The approach consists of five steps \cite{Camacho2019}, which are explained in what follows.

\textit{First}, raw network traffic data from different sources are parsed and transformed into a set of quantitative features, often using the feature-as-a-counter approach. Examples of such features are the number of times an event takes place, the count of a given word in a log, the number of times a given event takes place in a given time window, or the number of traffic flows with a given destination port in a NetFlow. The selection of the specific features and their parsing step require an effective comprehension of the data. \\
\indent \textit{Second}, all features are fused for the multivariate analysis, maintaining a common sampling rate. As a result, traffic flow matrices of $N$ observations featured in M-dimensional vectors $N\times M$ are ready to be analyzed.\\
\indent The \textit{third} and main step of MSNM is the anomaly detection, which is based on PCA. The well-known technique is applied to the mean-centered and auto-scaled $M$-dimensional dataset of $N$ observations ($ N \times M$), and projected into a subspace of range $P < M$ that maximizes the variance. To do so, original features are transformed into Principal Components (PC) using the eigenvectors of the covariance matrix $\bX^T\bX/N$. As a result, a residual matrix $\bE$ is generated as the differential error between projections and real samples. The transformation follows Eq. (\ref{eqPCA}), with $\bT$ ($N \times P$) and $\bV$ ($M \times P$) being the score and loading matrices, respectively :
\begin{equation}
    X=T\cdot V^T+E
    \label{eqPCA}
\end{equation}

Based on such transformation, MSNM proposes the usage of two complementary statistics extracted from the PCA analysis: 
\begin{itemize}
    \item The $Q$-statistic, also called \textit{Squared Prediction Error}, comprises the residuals in the n-th row of $\bE$ for a given observation $x_n$, following expression (\ref{estaQ}). As mentioned in Section~\ref{sec:intro}, $Q$ evaluates the reconstruction error of the projection used:
    \begin{equation}
        Q_n=e_n e_n^t
        \label{estaQ} 
    \end{equation}

    \item The $D$-statistic, or \textit{Hotelling's T2 statistic}, is computed by applying Eq. (\ref{estaD}) to PCA scores. As mentioned in Section I, $D$ represents a regularization term that rates how close the observation is to the data prior distribution. For a given observation $x_n$, and as $t_{n}$ is the score vector in the n-th row of $\bT$: 
    \begin{equation}
        D_n=t_n\Lambda^{-1}t_n^t
        \label{estaD}
    \end{equation}
\end{itemize}

Intuitively, $Q$ measures the capability of the model to recover a certain point in the data, while the regularization term $D$ measures the similarity of the latent representation with respect to those in the calibration data. As each term focuses on different domains, they are able to capture different types of anomalies.

It is well known that the $Q$-statistic has a high anomaly detection capability, and its usage together with the $D$-statistic is a key feature of the MSNM approach that offers attractive improvements \cite{macia2016hierarchical}. Once $D$ and $Q$ are calculated, they are used to model the normal operating conditions for the calibration of the MSNM system. In order to do so, \textit{upper control limits} ($U\!C\!L$) for a given significance level are defined for both $Q$, termed $U\!C\!L_Q$, and $D$, termed $U\!C\!L_D$. Diverse combinations of the two statistics can be used to provide a final expression for the anomaly evaluation. The authors of \cite{camacho2017traffic} propose the following weighting of $Q$ and $D$ to generate the anomaly score for a given observation $x_n$:
\begin{equation}
    t-Score_n=\frac{P\cdot D_n}{M \cdot U\!C\!L_D}+\frac{(M-P)\cdot Q_n}{M\cdot U\!C\!L_Q}
    \label{MSNM_score}
\end{equation}
\indent Once the anomaly is detected, the \textit{fourth} step of MSNM approximation is the pre-diagnosis. Features associated with the anomalies are identified in order to make an initial guess on their root causes. Contribution plots or others tools like oMEDA \cite{macia2016hierarchical} are commonly used to identify such features.\\
\indent Finally, the \textit{fifth} step consists of de-parsing the information pointed out during the detection (anomaly time-stamps) and pre-diagnosis (anomaly related features) phases. As a final result, raw information about the anomaly is extracted from specific logs or network traces. 

In what follows, we will introduce PPCA as a generative model framework (described in Section~\ref{sec:PPCA}) in order to derive the expressions for $Q$ and $D$ statistics, as well as the interpretation of Eq.~(\ref{MSNM_score}) in Section \ref{sec:PPCA_MSPC}.

\section{PPCA for anomaly detection}
\label{sec:PPCA}
Probabilistic PCA (PPCA) provides a method to calculate the principal subspace of a set of data vectors using a generative point of view and a maximum-likelihood framework. In order to be able to understand MSNM from a generative model perspective, it is important to first understand how PPCA is related to PCA.

This section provides a description of this connection, following the presentation of probabilistic PCA provided in \cite{bishop:2006}. While PCA is based on a deterministic linear projection of the data on a lower dimensional subspace, PPCA is a linear-Gaussian framework that considers a latent distribution for the data. Therefore a whole distribution of possible latent candidates is available for each observed data point. PPCA includes the measurement of deviations in the latent space, achieved in MSNM with the addition of the $D$ statistic, among other advantages (see \cite{tipping1999probabilistic} for a complete list).

Let $\bX$ denote the $N\times M$ network traffic matrix whose $i$-th row, $\bx_i^\T$, corresponds to the $i$-th observed instance, $i=1,\ldots,N$. That is, $\bX^\T=[\bx_1,\ldots,\bx_N]$. We assume that each column of $\bX^\T$ has been centered and normalized by its standard deviation. We also assume that calibrated observations are used.
For each instance (network traffic observation) an explicit latent variable $\bz$ with $P$ components is introduced. As we will see, it corresponds to components in a principal-component subspace. Next, a Gaussian prior distribution $\p(\bz)$ over the latent variable and a Gaussian observation distribution $\p(\bx|\bz)$ are introduced. Specifically:
\begin{align}
    \p(\bz)&={\cal N}(\bzero,\bI)  \label{eq:prior}\\
    \p(\bx|\bz)&={\cal N}(\bx|\bW\bz,\sigma^2\bI)\label{eq:observation}
\end{align}
where $\bW$ is a $M\times P$ matrix  whose columns span a linear (the principal-component) subspace within the data space, and the scale $\sigma^2$ governs the variance of the conditional distribution. Notice that, in what follows, we will omit the dependence of $\p(\bx|\bz)$ on $\bW$ and $\bsigma^2$ for simplicity. 

To estimate the values of the parameters $\bW$ and $\sigma^2$, we use maximum likelihood, and the marginal distribution $\p(\bx)$ is required. It can be easily calculated because the prior and the observation models in Eqs. (\ref{eq:prior}) and (\ref{eq:observation}) are both Gaussian. It follows that:
\begin{equation}
    \p(\bx)=\int \p(\bx|\bz)\p(\bz)\mbox{d}\bz={\cal N}(\bx|\b0,\bC),
\end{equation}
where 
\begin{equation}
\bC=\bW\bW^\T+\sigma^2\bI.
\end{equation}

The above likelihood requires the calculation of $\bC$ which may consume a lot of computational resources. This can be alleviated when $N$ is larger than $P$ (the dimension of the principal component subspace) and by utilizing the matrix inversion identity
\begin{equation}
    \bC^{-1}=\sigma^{-2}(\bI-\bW\bM^{-1}\bW^\T),
\end{equation}
where
\begin{equation}
    \bM=\bW^\T\bW+\sigma^2\bI.\label{eq:M}
\end{equation}

Using our normalized and calibrated observations, the maximum likelihood estimates are calculated by solving
\begin{align}
    &\bW_{\mbox{\small ML}},\sigma^2_{\mbox{\small ML}}=\arg\max_{\bW,\bsigma^2}\sum_{n=1}^N\ln \p(\bx_n|\bW,\bsigma^2) \nonumber \\
    &=\arg\max_{\bW,\bsigma^2} -\frac{NP}{2}\ln(2\pi)-\frac{N}{2}\ln|\bC|-\frac{1}{2}\sum_{n=1}^N\bx_n^\T\bC^{-1}\bx_n\label{eq:marginal} \nonumber \\
    &=\arg\max_{\bW,\bsigma^2} -\frac{N}{2}\{P\ln(2\pi)+\ln|\bC|+\mbox{tr}(\bC^{-1}\bS)\}
\end{align}
where $\bS$ is the data normalized and calibrated covariance matrix.

Maximizing the above equation with respect to $\bW$ and $\sigma^2$ is not easy. However, it can be shown ---see \cite{bishop:2006}--- that 
\begin{equation}
    \bW_{\mbox{\small ML}}=\bU(\bL-\sigma^2\bI)^{1/2}\bR \label{eq:W} 
\end{equation}
where $\bU$ is a $M\times P$ matrix whose columns correspond to the $P$ eigenvectors associated with the $P$ largest eigenvalues of the data normalized and calibrated covariance matrix $\bS$, $\lambda_1,\ldots,\lambda_P$. $\bL$ is a diagonal matrix with diagonal values these eigenvalues, and $\bR$ is an $P\times P$ orthonormal matrix that represents any rotation. Note that $\sigma^2$ is constrained to be smaller than the lowest eigenvalue $\lambda_P$ (the minimum element in the diagonal matrix $\bL$, thus avoiding a negative square root in Eq. (\ref{eq:W})). Thus, $\sigma^2\in[0,\lambda_P)$.  Furthermore, the maximum likelihood for $\sigma^2$ is 
\begin{equation}
    \sigma^2_{\mbox{\small ML}}=\frac{1}{M-P}\sum_{i=P+1}^M\lambda_i \label{eq:sigma} 
\end{equation}

Note that with the estimated $\bW_{\mbox{\small ML}}$ and $\sigma^2_{\mbox{\small ML}}$ we can calculate, for a given normalized sample $\bx$, the quantity
\begin{equation}
\ln \p(\bx|\bW_{\mbox{\small ML}},\bsigma_{\mbox{\small ML}}^2)=-\frac{M}{2}\ln(2\pi)-\frac{1}{2}\ln|\bC|-\frac{1}{2}\bx^\T\bC^{-1}\bx \label{eq:mlikelihood}
\end{equation} 
and use it to decide whether $\bx$ is an anomaly (we have omitted the dependency of $\bC$ on $\bW_{\mbox{\small ML}}\mbox{ and }\bsigma_{\mbox{\small ML}}^2$ for simplicity). The whole process for anomaly detection in PPCA is represented in algorithm \ref{alg:PPCA}. First, the data need to be mean-centered around zero and scaled as PCA and PPCA are sensitive to feature scaling. Then, the parameters of the model are estimated using Eq. (\ref{eq:sigma}) and Eq (\ref{eq:W}) on trusted calibration data. Once the parameters are fixed, new data can be checked for anomalies.

\begin{algorithm}[h]
\caption{PPCA for anomaly detection.\label{alg:PPCA}}
\begin{algorithmic}
\REQUIRE Centered and normalized observations $\bX$, predefined threshold $\mbox{\textit{thr}}$.
\STATE Obtain $\sigma^2_{\mbox{\small ML}}$ using Eq. (\ref{eq:sigma}).
\STATE Obtain $\bW_{\mbox{\small ML}}$ using Eq. (\ref{eq:W}).
\STATE For a new sample $\bx^{new}$, decide whether it is an anomaly by thresholding $\ln \p(\bx^{new}|\bW_{\mbox{ML}},\bsigma_{\mbox{\small ML}}^2)$ defined in Eq. (\ref{eq:mlikelihood}), that is if,  
\begin{equation}
    \frac{1}{2}\bx^\T\bC^{-1}\bx>\mbox{\textit{thr}}.
\end{equation}
\end{algorithmic}
\end{algorithm}




Notice that Algorithm \ref{alg:PPCA} can be used even with a value $\sigma^2$ that is different from $\sigma^2_{\mbox{ML}}$. The only constraint on $\sigma^2$ is that it has to be smaller than $\lambda_P$, see Eq. (\ref{eq:W}).

Using $\frac{1}{2}\bx^\T\bC^{-1}\bx>\mbox{\textit{thr}}$ to detect anomalies does not provide an insight into the roles played by the prior and the observation models in the detection process. This is crucial since it will allow us to relate PPCA and MSNM, as we will see in the following section.

\section{Relating PPCA to MSNM}
\label{sec:PPCA_MSPC}
\subsection{Revisiting PPCA for anomaly detection}
\label{sec: relating A}
In order to relate PPCA and MSNM we will evaluate other variance values apart from 
$\sigma_{\mbox{\small ML}}^2$, in the expressions for PPCA. Thus, for clarity, we will make the formulations using a generic variance $\delta$ (smaller than $\lambda_P$). We will use the Laplace approximation \cite{bishop:2006} described in Appendix A  as a starting point for the analysis. Given a marginal probability $\bp(\bx|\bW,\delta)$, with $\bW$ and $\delta$ being its score matrix and variance, and following Eq. (\ref{app:marg}), it follows that:
\begin{equation}
    \ln \p(\bx|\bW,\delta)=-\frac{1}{2}(\hat\bz^\T\hat\bz+\frac{1}{\delta}\parallel \bx-\bW\hat\bz \parallel^2)+\mbox{const}.
    \label{app:marg1}
\end{equation}
where $\hat\bz$ is the mode of the posterior distribution $\p(\bz|\bx)$. 

Therefore, leaving constants aside for the sake of clarity, we can equalise the thresholding expression of $\ln \bp(x|\bW,\delta)$ for PPCA ---Eq. (\ref{eq:mlikelihood})--- to its Laplace approximation: 
\begin{equation}
    \frac{1}{2}\bx^\T\bC^{-1}\bx= \frac{1}{2}(\hat\bz^\T\hat\bz+\frac{1}{\delta}\parallel \bx-\bW\hat\bz \parallel^2)\label{eq:split_marginal}
\end{equation}
In order to calculate the term $\hat{\bz}$ in Eq. (\ref{eq:split_marginal}), the closed expression for $\bp(z|x)$ derived in \cite{bishop:2006} can be used:
\begin{equation}
    \p(\bz|\bx)={\cal N}(\bz|\bM^{-1}\bW^\T\bx,\delta\bM^{-1})
\end{equation}
and since for this Gaussian distribution the mode and the mean coincide, it follows that:
\begin{equation}
    \hat\bz=\bM^{-1}\bW^\T\bx
    \label{eq:moda}
\end{equation}

As seen in Eq.~(\ref{eq:W}) (Section \ref{sec:PPCA}), $\bW$ depends on the variance $\delta\in[0,\lambda_P)$:
\begin{equation}
    \bW(\delta)=\bU(\bL-\delta\bI)^{1/2}, \label{eq:W1} 
\end{equation}
 
\noindent where for the sake of simplicity, $\bR=\bI$ is used in \eqref{eq:W}. Introducing Eq. (\ref{eq:W1}) in Eq.~(\ref{eq:M}), and taking into account that $\bU^\T\bU = \bI$, it follows that:
 \begin{equation}
\bM=\bW(\delta)^\T\bW(\delta)+\delta\bI=(\bL-\delta\bI)+\delta\bI=\bL\label{eq:M1}
\end{equation}

Using Eqs. (\ref{eq:W1}) and (\ref{eq:M1})  in Eq. (\ref{eq:moda}):
\begin{equation}
\hat\bz(\delta)=\bM^{-1}\bW(\delta)^\T\bx=\bL^{-1}(\bL-\delta\bI)^{1/2}\bU^\T\bx.\label{eq:map1}
\end{equation}

In short, the whole process for anomaly detection with this revisited PPCA is summarized in Algorithm \ref{alg:PPCA:sum}. Note that it is equivalent to Algorithm \ref{alg:PPCA}, and allows arbitrary variance values to be used. The parameters of the model are estimated on trusted calibration data before checking new data for anomalies.

\begin{algorithm}[h]
\caption{PPCA for anomaly detection (Revisited) \label{alg:PPCA:sum}}
\begin{algorithmic}
\REQUIRE Centered and normalized observations $\bX$, predefined threshold $\mbox{\textit{thr}}$.
\STATE Set a certain variance $\delta\in[0,\lambda_P)$.
\STATE Obtain $\bW(\delta)$ using Eq. (\ref{eq:W1}).
\STATE Obtain $\bM(\delta)$ using Eq. (\ref{eq:M1}).
\STATE Obtain $\hat\bz(\delta)$ using Eq. (\ref{eq:map1}).
\STATE For a new sample $\bx^{new}$, decide whether it is an anomaly by thresholding $\ln \p(\bx^{new}|\bW(\delta),\delta)$ defined in Eq. (\ref{app:marg1}), that is if, 
\begin{align}
\frac{1}{2}(\hat\bz^\T(\delta)\hat\bz(\delta)
&+\frac{1}{\delta}\parallel \bx-\bW(\delta)\hat\bz(\delta) \parallel^2)\nonumber\\
>\mbox{\textit{thr}}
\end{align}
\end{algorithmic}
\end{algorithm}

\subsection{Analysis of the variance: connecting PPCA and MSNM}
Recall that Algorithm \ref{alg:PPCA:sum} can be used with a value of $\delta\in[0,\lambda_P)$ which is different from $\sigma_{\mbox{\small ML}}^2$. We can observe the connection between PPCA and MSNM when studying the range of values that this variance $\delta$ might take. To do so, let us analyze the expression of ln $ \bp(\bx|W,\delta)$ in Eq. (\ref{app:marg1}). Leaving aside constant terms, Eq. (\ref{app:marg1}) is formed using two terms, the first of which considers the influence of the latent space in the probability calculation, and somehow acts as a \textit{regularization term}:

\begin {equation}
\label{eq:regularizationterm}
\hat{z}(\delta)^T\hat{z}(\delta)
\end{equation}
 while the second evaluates the difference between a sample $\bx$ and its reconstruction from the latent space. Thus, it plays the role of a \textit{reconstruction error}:
 \begin{equation}
\parallel x-W\hat{z}(\delta)\parallel^2
\label{eq:error}
\end{equation}
Note that the reconstruction error contribution to the sample's probability is weighted by a factor $1/\delta$. At this point, we separate the influence of $\delta$ in this weighting factor from that of $\delta$ in $\bW(\delta)$ and $\hat{z}(\delta)$. Thus, we let the weighting factor take a fixed value $1/\alpha$ while $\delta$ keeps taking possible values in $[0,\lambda_P)$. Thus, a function $f_{\alpha}(\delta)$ can be defined to study the behavior of  ln $\p(\bx|W,\delta)$ for different values $\delta \in [0,\lambda_P)$ :
\begin{align}
\label{eq:PPCA_MSPC_relation}
f_{\alpha}(\delta)=\frac{1}{2}(\hat\bz^\T(\delta))\hat\bz(\delta)+\frac{1}{\alpha}\parallel \bx-\bW(\delta)\hat\bz(\delta) \parallel^2)
\end{align} 
It can be seen that ---see Appendix B and Figure~\ref{fig:functionAnalysis}---  $f_{\alpha}(\delta)$ in Eq. (\ref{eq:PPCA_MSPC_relation}) has 3 properties:
\begin{itemize}
    \item [(i)] $f_{\alpha}(\delta)$ is convex,
    \item [(ii)] its minimum value is achieved at $\delta=\alpha/2$,
    \item [(iii)] $f_{\alpha}(0)=f_{\alpha}(\alpha)$.
\end{itemize}
Based on the fact that MSNM uses PCA for its modelling and PPCA converges to PCA when $\delta=0$ --- see \cite{bishop:2006}---, we have explored the value of $f_{\alpha}(\delta)$ for $\delta = 0$: 
\begin{align}
f_{\alpha}(0)&=\frac{1}{2}(\bx^\T\bU\bL^{-1}\bU^\T\bx +\frac{1}{\alpha}\parallel\bx-\bU\bU^\T\bx\parallel^2) 
\label{eqn:cero}
\end{align}
Eq. (\ref{eqn:cero}) is exactly the quantity used by MSNM to detect anomalies since:
\begin{itemize}
    \item $\bx^T \bU \bL^{-1}\bU^\T\bx$, the regularization part of the probability, matches Eq. (\ref{estaD}) that defines the $D$-statistic.
    \item $\parallel\bx-\bU\bU^\T\bx\parallel^2$, the reconstruction error part of the probability, matches Eq. (\ref{estaQ}) that defines the $Q$-statistic. 
    \item Both terms, regularization and reconstruction error, contribute with a different weighting: \textit{i)} $1/\sigma_{\mbox{\small ML}}^2$ in PPCA, and \textit{ii)} the empirical value  $(M-P/M)$ ---see Eq. (\ref{MSNM_score})--- in MSNM.
\end{itemize} 

The results obtained show that the conditions that make MSNM and PPCA coincide are:

\begin{align}
    \label{eq:conditionEqualMSNMPPCA}
    \alpha &= \sigma^2_{\mbox{\small ML}} \\
    \delta &=  \sigma^2_{\mbox{\small ML}} \mbox{ or } \delta=0\nonumber
\end{align} 

The behavior of the function $f_{\alpha}(\delta)$ is shown in Figure~\ref{fig:functionAnalysis}. 

\begin{figure}
\centering
    \includegraphics[width=0.7\columnwidth]{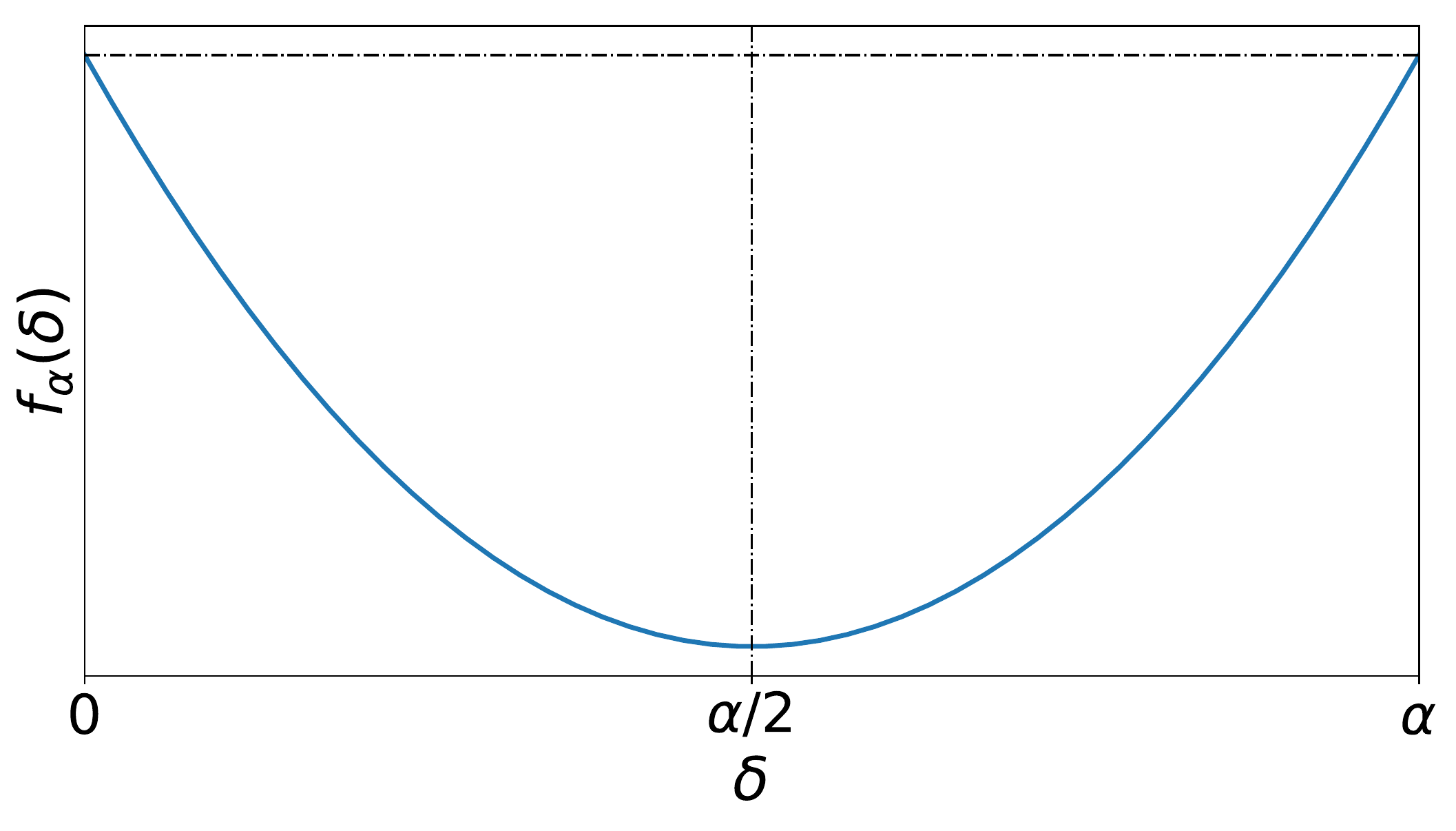}
    \caption{Analysis of $f_{\alpha}(\delta)$.}
    \label{fig:functionAnalysis}
\end{figure}

Notice also that in this \textit{revisited version of PPCA}  we could use a value of $\delta$ that is different from zero and $\sigma_{\mbox{\small ML}}^2$. In that case, for a given anomaly probability threshold, 
the anomaly detection model would accept less observations (lower probability) if $\delta\in [0,\alpha]$ and more if $\alpha<\delta<\lambda_P$.

To end Section~\ref{sec:PPCA_MSPC}, note that the conclusions obtained for PPCA are applicable to other generative models based on PPCA, like Variational Autoencoders (VAEs). In VAEs, the calculation of $\p(\bz|\bx)$ in a closed form is not possible and, thus, we must rely on the so called Evidence Lower Bound (ELBO), which comprises two different terms (as in PPCA): a reconstruction error and a regularization term. Our conclusions for PPCA lead to the same recommendations for VAEs, \textit{i.e.}, both terms should be used in the calculation of anomaly scores. We consider this to be a relevant contribution of this paper, as we can still find examples in the state-of-the-art  \cite{Xu2018VAE}\cite{LopezMartin2017CVAE}\cite{An2015VAE}\cite{Zavrak2020VAE} that only use the error reconstruction term for anomaly detection. 


\section{Experimental Validation}
\label{sec:experiments}
\subsection{Datasets Description}
\label{sec:datasets}

To experimentally validate the theory introduced in the previous sections, we will make use of two datasets: one with synthetic data, and another with real network traffic information, as described below. 

\subsubsection{Synthetic dataset}
\begin{figure}
    \centering
    \includegraphics[width=0.48\columnwidth]{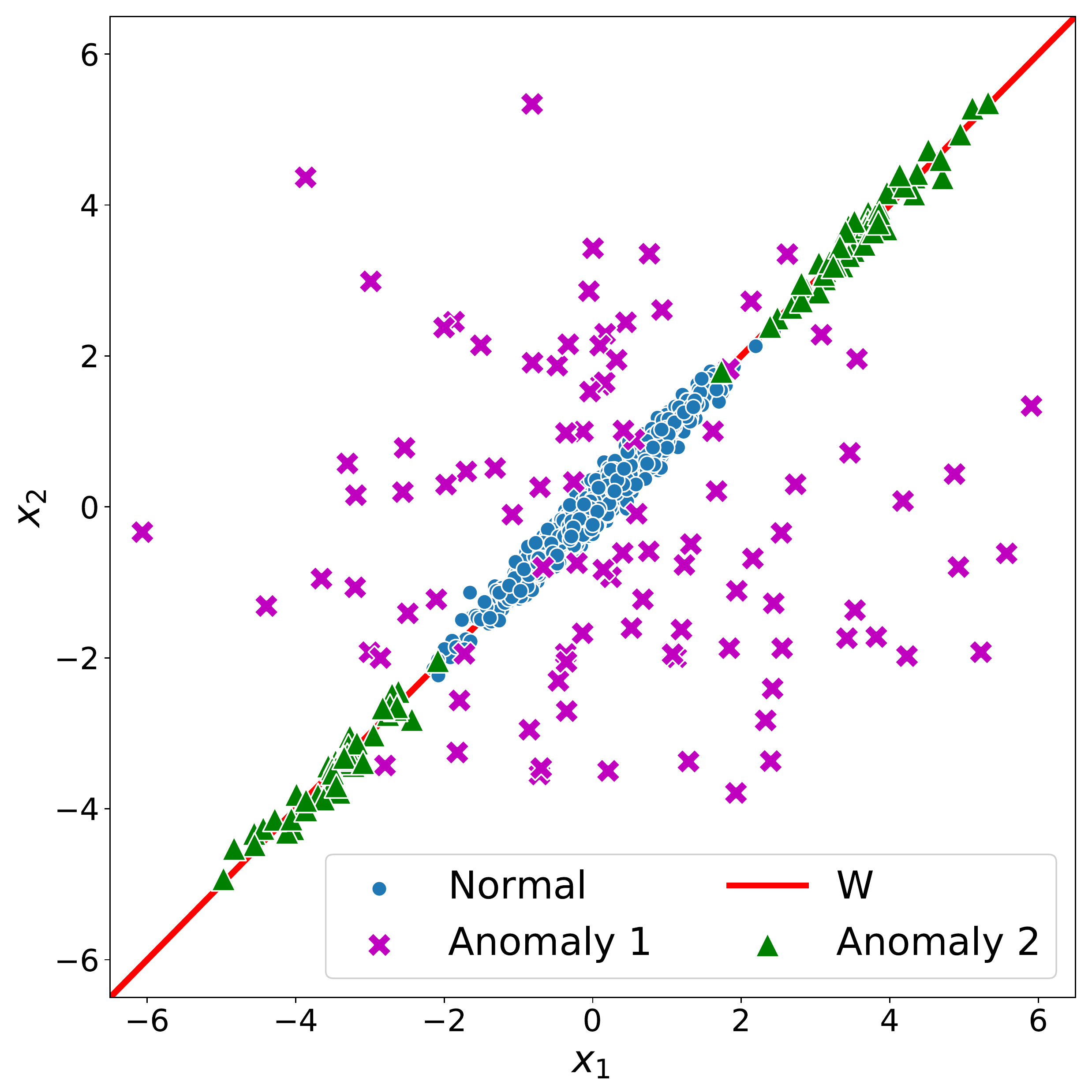}
    \caption{Synthetic dataset test bench. Normal data is presented in blue circles, Anomaly type 1 is presented in fuchsia with \textbf{x}-shaped markers, while the green triangles refer to Anomaly type 2. The arbitrary line $\bW$ used to create the data is plotted in red.}
    \label{fig:Synthetic_dataset}
\end{figure}

The synthetic dataset is intended to provide a 2-dimensional plottable scenario that provides an easy interpretation of the anomalies and behaviour of the different models. Using the PCA observation model given by $\bx =\bW\bz +\epsilon,$ we obtain $N$ samples of $\bz$ from a Gaussian distribution with zero mean and unit variance. Then we multiply $\bz$ by an arbitrary $\bW=[0.707,0.707]^T$ and add $\epsilon$ sampled from ${\cal N}(\bzero,0.1\cdot \bI)$ to obtain N bidimensional samples of $\bx$. The samples will follow a Gaussian distribution along the selected $\bW$. Using this method, we generate 1000 samples of clean data to calibrate our models. For the testing set, we mix 1000 new clean data samples with two different types of anomalies. The first type of anomaly is sampled from a distribution that is different from that of the above described samples. Specifically, we choose a multivariate Gaussian with zero mean and a diagonal variance matrix $5\times\bI$. Anomaly type~1 is not designed according to the linear model. The second type of anomaly is generated in the latent space, we sample $z_{anom}$ using a Gaussian with mean $5$ and unit variance. The values of $z_{anom}$ are then randomly multiplied by $-1$ following a Bernoulli (0.5) distribution. This procedure provides anomalies that follow the generative model but whose latent distribution is different from that of the calibration data. For each type of anomaly mentioned, 100 data points are introduced. See Figure~\ref{fig:Synthetic_dataset} for a visual representation of the data and anomalies. 

\subsubsection{UGR'16 dataset}
The UGR'16 dataset~\cite{MACIAFERNANDEZ_UGR16} was designed for the evaluation of cyclostationary-based network IDSs. It contains real anonymized NetFlow traces captured over several months in a tier-3 ISP. The traces include legitimate traffic from many virtualized services in the cloud, like web servers with proprietary and standard configurations, and other hosted services like DNS, FTP, mail servers, etc. In addition, a set of malicious virtual machines was configured in the network to generate attack traffic. The traffic is captured from two border routers of the ISP network. Thus, the dataset includes both legitimate traffic and realistic attack scenarios, all of them labeled. 

UGR'16 divides the data in two differentiated sets: calibration and test. The calibration set contains real background traffic data gathered from March to June in 2016 (4 months). The test set mixes real background traffic and synthetically generated malicious traffic, gathered from July to August 2016. Although the data was captured in 2016, it was published in 2018 and is still considered of interest in recent work \cite{Ring_survey, Castillo_DoSDatasets}.

To train our models, we use the data gathered during working days in May, where less anomalies were detected after data obtained was analyzed \cite{MACIAFERNANDEZ_UGR16}, and no synthetically generated attacks were introduced. The test set uses data gathered on those working days when synthetically generated attacks were interlaced within background traffic. The attack types that were synthetically generated include:
\begin{itemize}
    \item Low-rate Denial of Service (DoS): TCP SYN packets are sent to the victims port 80 using 1280-bit packets and a rate of 100 packets/s. The rate is not high enough to avoid the normal operation of the network being affected.
    \item Port scanning: A continuous SYN scan of common ports of victims. Two variants are implemented for this attack: Scan11 (one-to-one scan attack) and Scan44 (Four-to-four scan attack).
    \item Botnet traffic: Obtained from the execution of the malware known as Neris in a controlled environment (See \cite{GARCIA2014Botnet} for details about the malware and \cite{MACIAFERNANDEZ_UGR16} for details about its injection in the data.).
\end{itemize}

\begin{table}[]
    \centering
    \caption{Summary of data flows from UGR'16 used to train and test the models}
    \begin{tabular}{ccc}
    \toprule
        Flow type & Calibration & Test \\
        \midrule
        Background traffic & 31680 & 8714 \\
        DoS &NA& 299\\
        Scan44 &NA& 65 \\
        Scan11 &NA& 66\\
        Nerisbotnet &NA& 488 \\
        UDPscan &NA& 9\\
        SSHscan &NA& 9\\
        Spam &NA& 3616\\
        \bottomrule
    \end{tabular}

    \label{tab:UGR16_overview}
\end{table}

The test set also included labels for a real UDP Scan campaign that was identified in the background traffic. 

The labels are provided as a list of timestamps (in mins) of when the attacks were executed. Table \ref{tab:UGR16_overview} summarizes the traces in the train and test sets. The NetFlow logs cannot be directly used to feed PCA-based anomaly detection systems \cite{lakhina2004characterization}. Thus,  following \cite{Camacho2019}, we use the FCParser\footnote{Available at: https://github.com/josecamachop/FCParser} tool to extract 143 quantitative features from the NetFlow logs. Each feature counts the number of times that a given event takes place during each minute \emph{, e.g.,} number of flows with a given destination port, number of flows with an accumulated payload size greater than a threshold, etc. Features were manually defined in \cite{Camacho2019} from domain knowledge using regular expressions. Therefore, they represent information that experts would use to manually identify anomalies in the data. In this paper, we focus on differentiating the minutes labeled as anomalies from those labeled as background data. We study each type of anomaly/attack separately. The anomaly scores are calculated for the whole test set at once. Then, classification metrics are calculated for each anomaly type against background data only (binary classification) 

\begin{figure}
    \centering
    \includegraphics[width=\columnwidth]{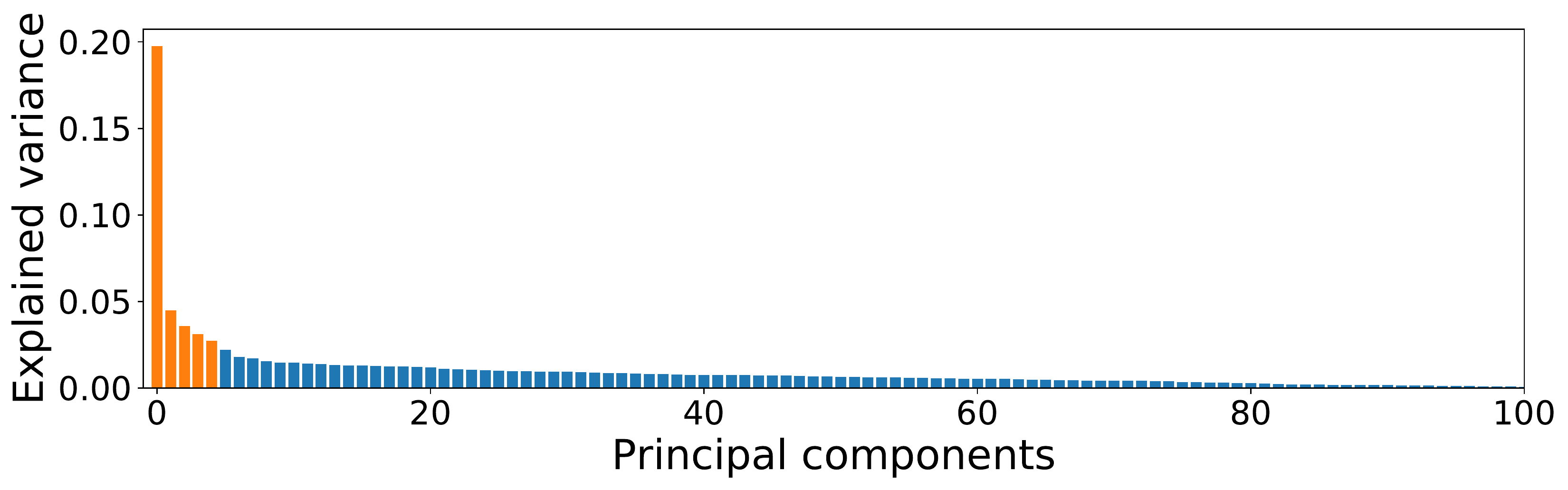}
    \caption{Explained variance per component in the UGR'16 dataset.}
    \label{fig:Explained_Var_UGR16}
\end{figure}

The evaluation of the variance captured  by the different principal components of PCA on the calibration set (see Figure~\ref{fig:Explained_Var_UGR16}), shows that the principal components 1-5 are the most relevant and capture a higher percentage of variance than the rest. The model with 5 P explains 33.65\% of the total variance.  

\subsection{Experiments}
This section focuses on describing three different experiments: \textit{i)} Anomaly detection using the MSNM and PPCA models with a common threshold to show that detection results from both models coincide. \textit{ii)} Anomaly detection using only either the regularization or the reconstruction error term separately, to analyze their performance for different types of anomalies. \textit{iii)} Finally, we assess the correctness of the weighting values for both regularization and reconstruction error terms, for MSNM and PPCA models. 


\subsubsection{MSNM and PPCA equivalence}
\label{sec:MSNMvsPPCA}

To experimentally demonstrate the MSNM and PPCA equivalency shown in Section~\ref{sec:PPCA_MSPC}, we perform 
anomaly detection on the datasets presented in Section~\ref{sec:datasets} using both MSNM and PPCA models. 
The anomaly score is calculated using Eq.~\eqref{eq:PPCA_MSPC_relation} with the condition in Eq.~(\ref{eq:conditionEqualMSNMPPCA}), \textit{i.e.},   $\alpha=\sigma^2_{\mbox{\small ML}}$ and $\delta=0$ or $\delta=\sigma^2_{\mbox{\small ML}}$ for the MSNM and PPCA models, respectively. When a threshold value is needed, it should be chosen in accordance with the values in the calibration set, according to our confidence in the absence of malicious traffic in the training data. In our case, the threshold is the 99-percentile of the anomaly scores of the calibration set for each model. The threshold can be modified if the confidence in the calibration data is reduced.

\begin{table}[]
    \centering
    \caption{Performance metrics for the synthetic dataset using different threshold values selected according to different confidence levels on the calibration set.}
    \begin{tabular}{cccccc}
    \toprule
         Confidence level&95\%&96\%&97\%&98\%&99\%  \\
         \midrule
         Threshold $\chi^2$&5.99&6.43&7.01&7.82&9.21 \\
         Accuracy&0.9925&0.9933&0.9925&0.9866&0.9808 \\
         False Alarm Ratio&0.002&0.001&0.001&0.001&0 \\
    \bottomrule
    \end{tabular}
    
    \label{tab:Threshold_values}
\end{table}

Note also that once $\sigma^2_{\mbox{\small ML}}$ and $\bW_{\mbox{\small ML}}$ have been calculated, Eq. \eqref{eq:mlikelihood} can be used to set a non-experimental value. Since $p(x|\bW_{\mbox{\small ML}},\sigma^2_{\mbox{\small ML}})$ is a normal distribution we can select a threshold $\delta$ such as 

\begin{equation}
    \int_{\bx^\T\bC^{-1}\bx\le \delta}p(x|\bW_{\mbox{\small ML}},\sigma^2_{\mbox{\small ML}})d\bx \ge \alpha
\end{equation}
for a confidence level $\alpha$. We can take into account that $\bx^\T\bC^{-1}\bx\sim \chi^2(M)$ and select for this distribution a threshold $\delta$ with a given confidence level $\alpha$. This theoretical approach is less used in practice.

\begin{figure}
    \centering
    \includegraphics[width=0.48\columnwidth]{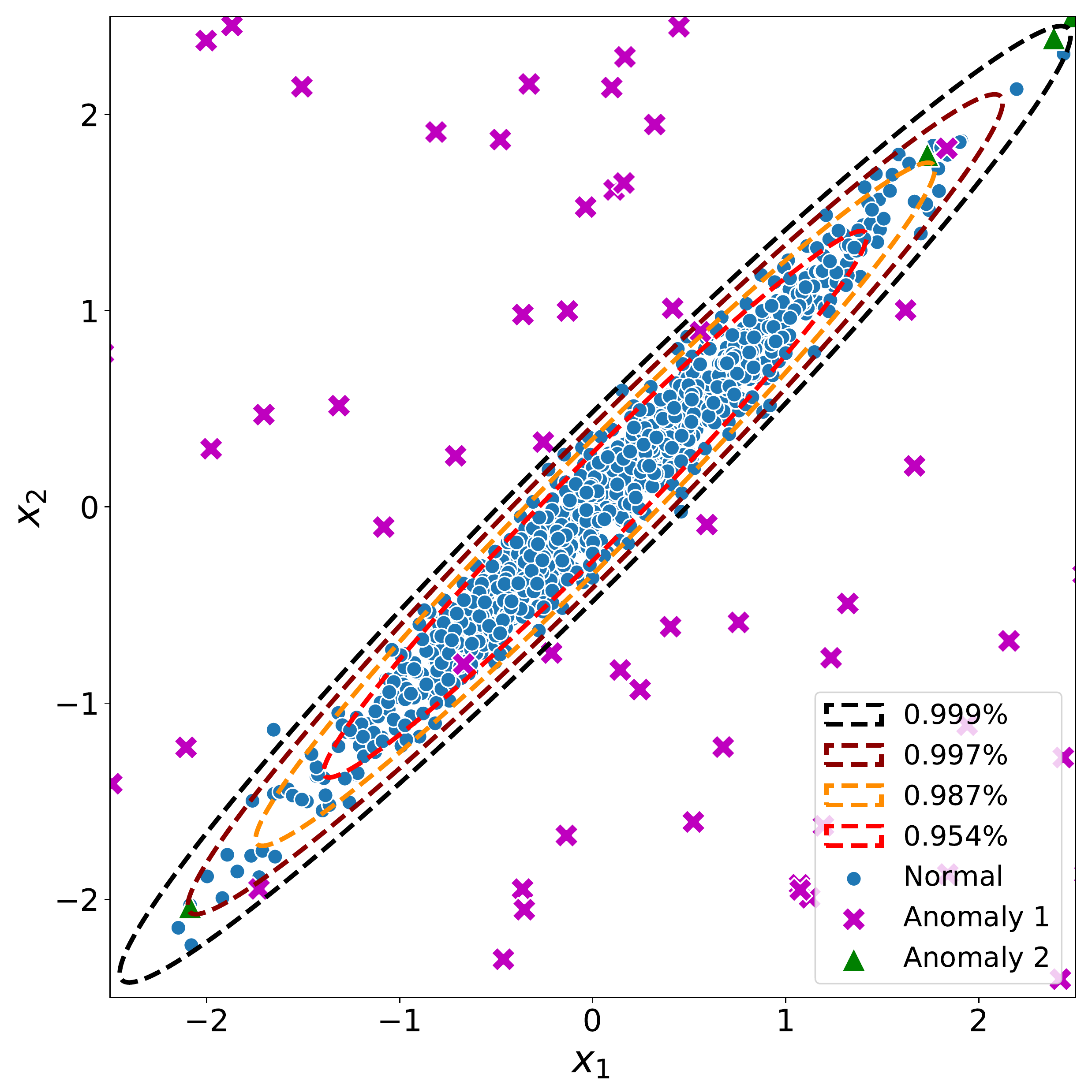}
    \caption{Detail of the synthetic test set shown in Figure \ref{fig:Synthetic_dataset}. The detection area obtained with the threshold for different confidence levels $\alpha$ on the calibration data is shown in dashed lines .}
    \label{fig:Threshold_values}
\end{figure}

In Figure~\ref{fig:Threshold_values} we can see how the detection area changes according to the confidence level $\alpha$. Table \ref{tab:Threshold_values} includes the threshold value, accuracy and False Alarm Ratio using the theoretical bound $\bx^\T\bC^{-1}\bx\le \delta$.

\begin{figure*}[t]
    \centering
     \setlength{\tabcolsep}{1.7pt} 
\setlength\fboxsep{0pt}
\pdfpxdimen=\dimexpr 1 in/72\relax

    \begin{tabular}{ccc}
    \includegraphics[width=0.33\textwidth]{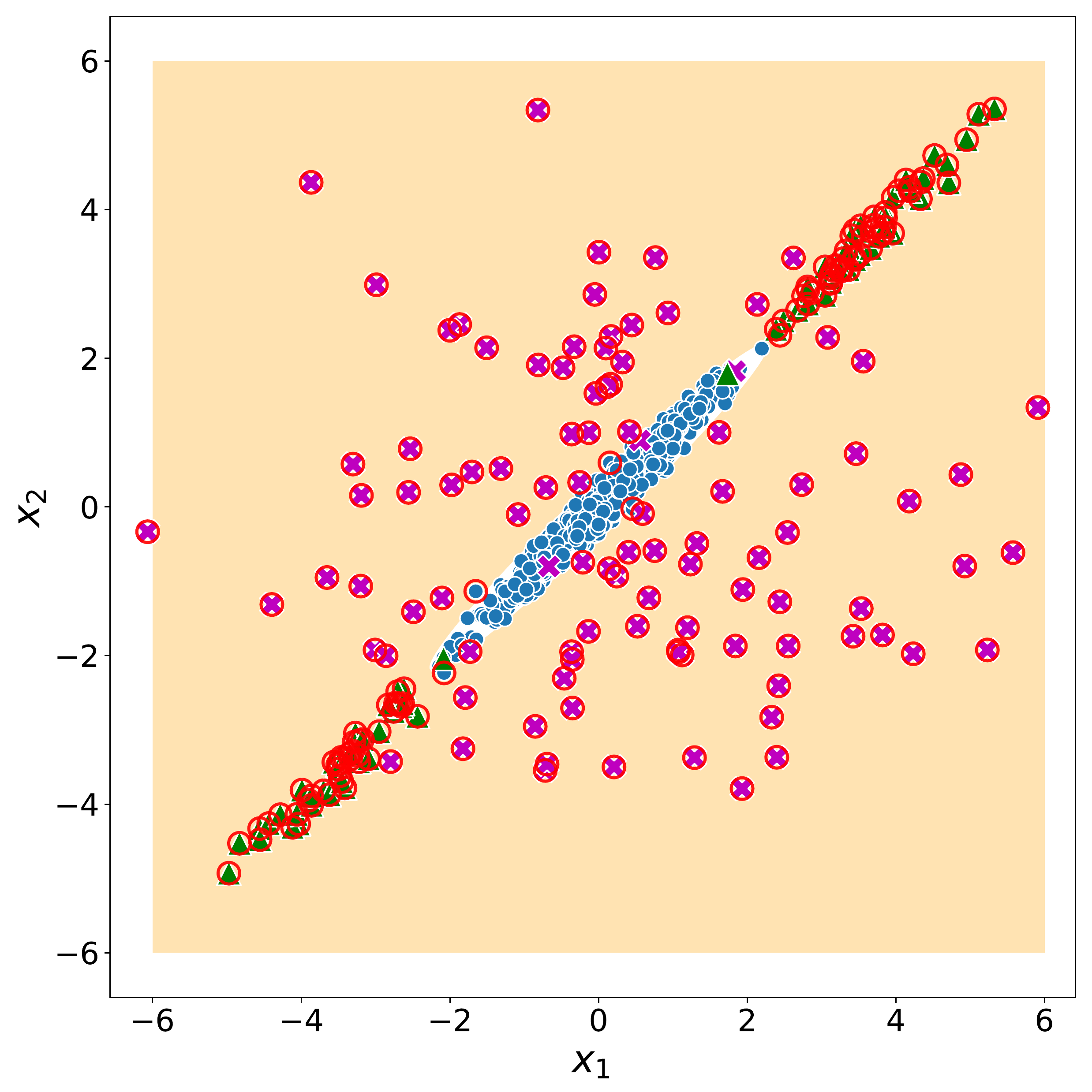}
    &
    \includegraphics[width=0.33\textwidth]{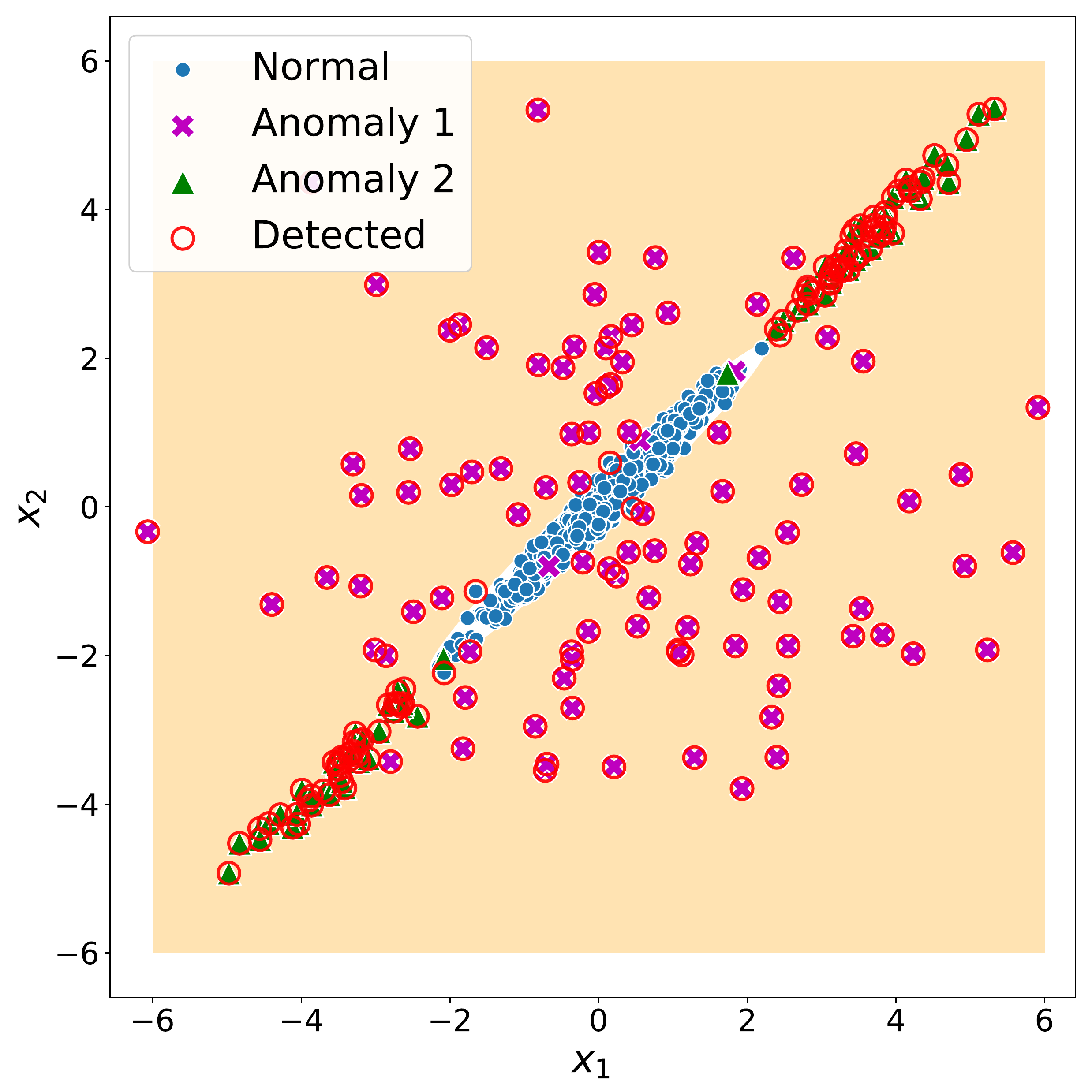}
    &
    \includegraphics[width=0.33\textwidth]{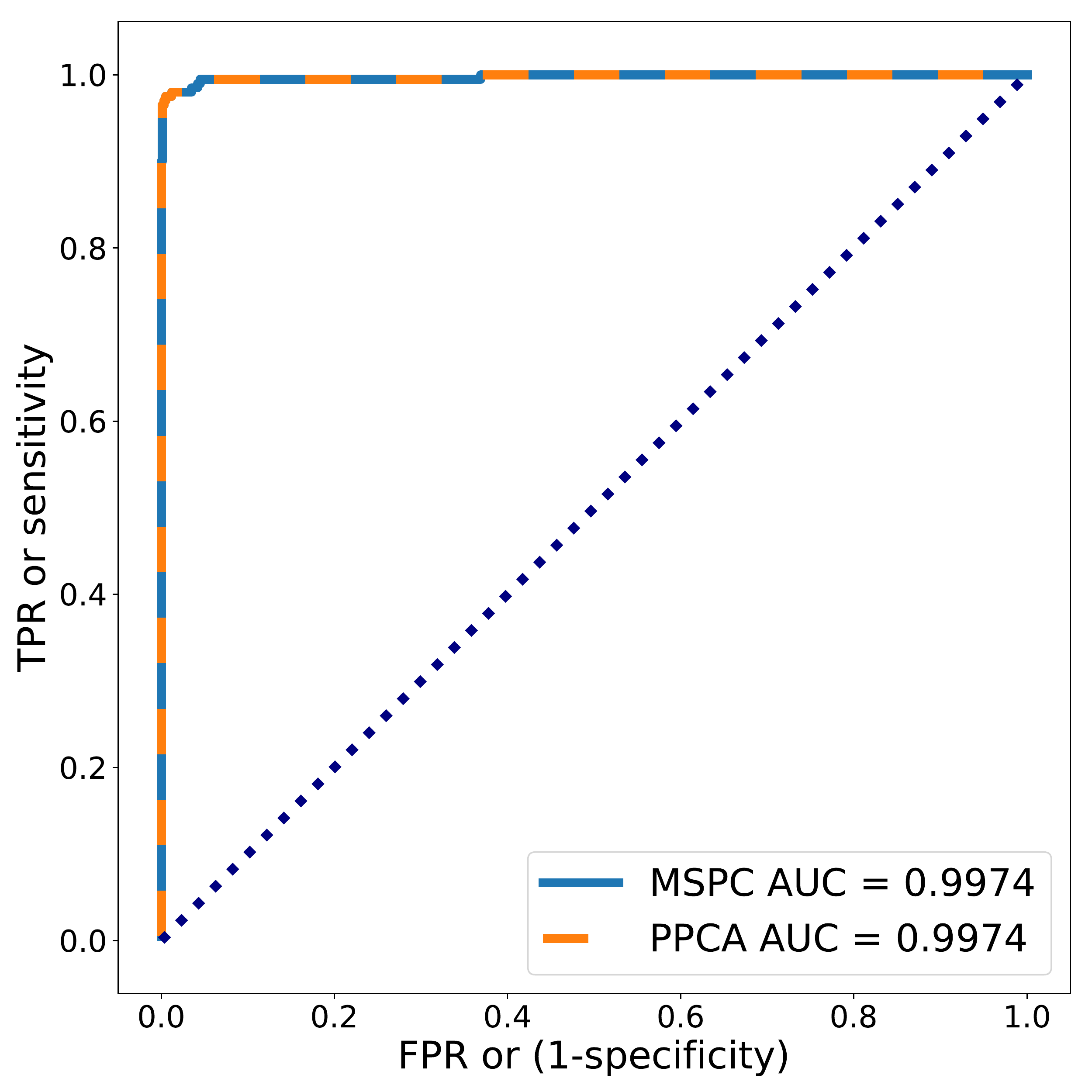}
    \\
a) MSNM   & b) PPCA & c) ROC curves for both models\\

    \end{tabular}
    \centering
    \caption{Anomaly detection on the synthetic dataset for both \textit{a)} MSNM and \textit{b)} PPCA models. Blue circles identify normal data, fuchsia \textbf{x}s and green triangles identify different types of anomalies. Red circumferences mark the points detected as anomalies. The orange shadows cover the detection area in each case. \textit{c)} ROC curves for both models. }
    \label{fig:Toy_MSPC_PPCA_comparison}
\end{figure*}

\begin{figure}[t]
    \centering
     \setlength{\tabcolsep}{1.7pt} 
\setlength\fboxsep{0pt}
\pdfpxdimen=\dimexpr 1 in/72\relax

    \begin{tabular}{cc}
    \includegraphics[width=0.48\columnwidth]{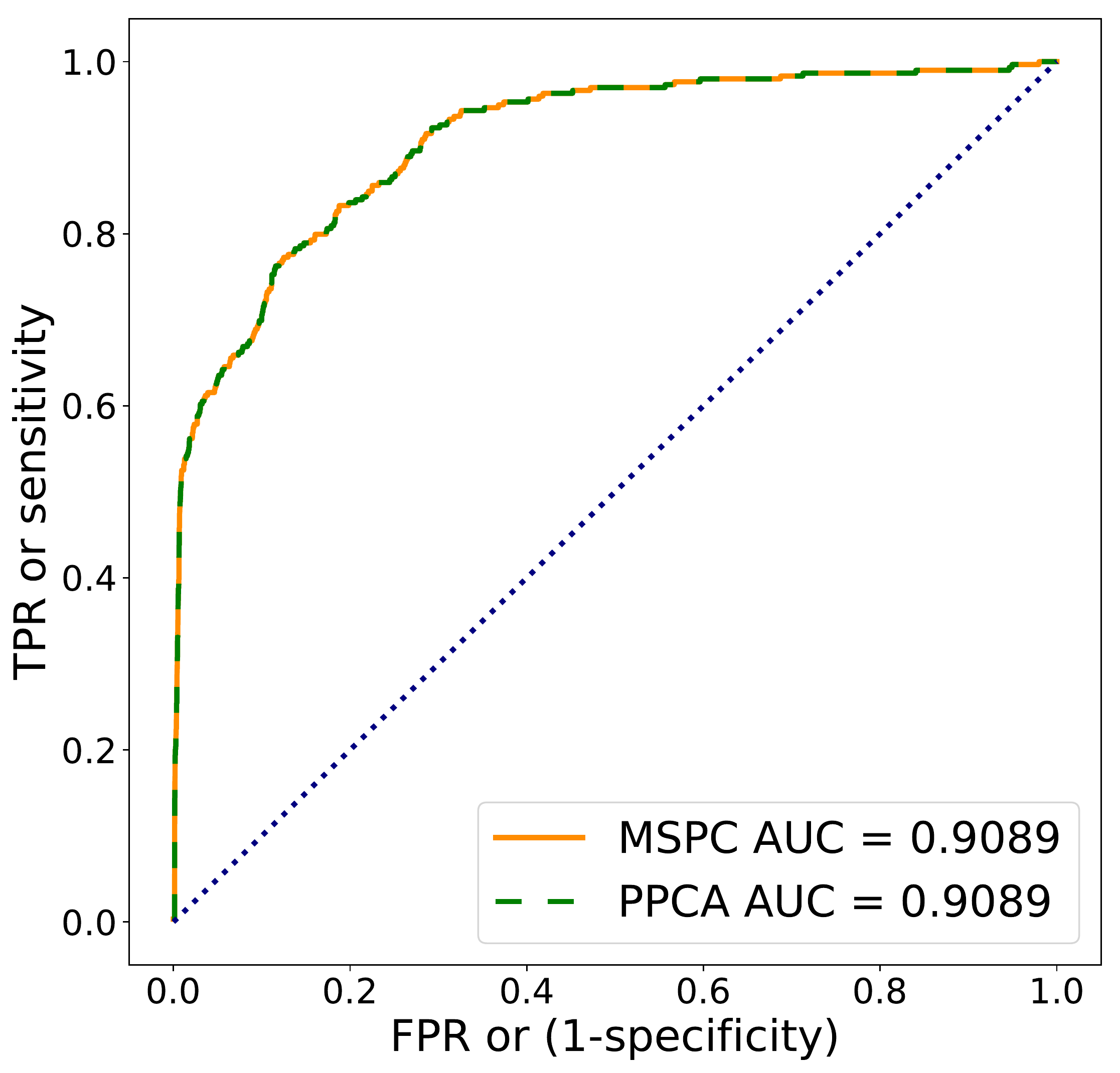}
    &
    \includegraphics[width=0.48\columnwidth]{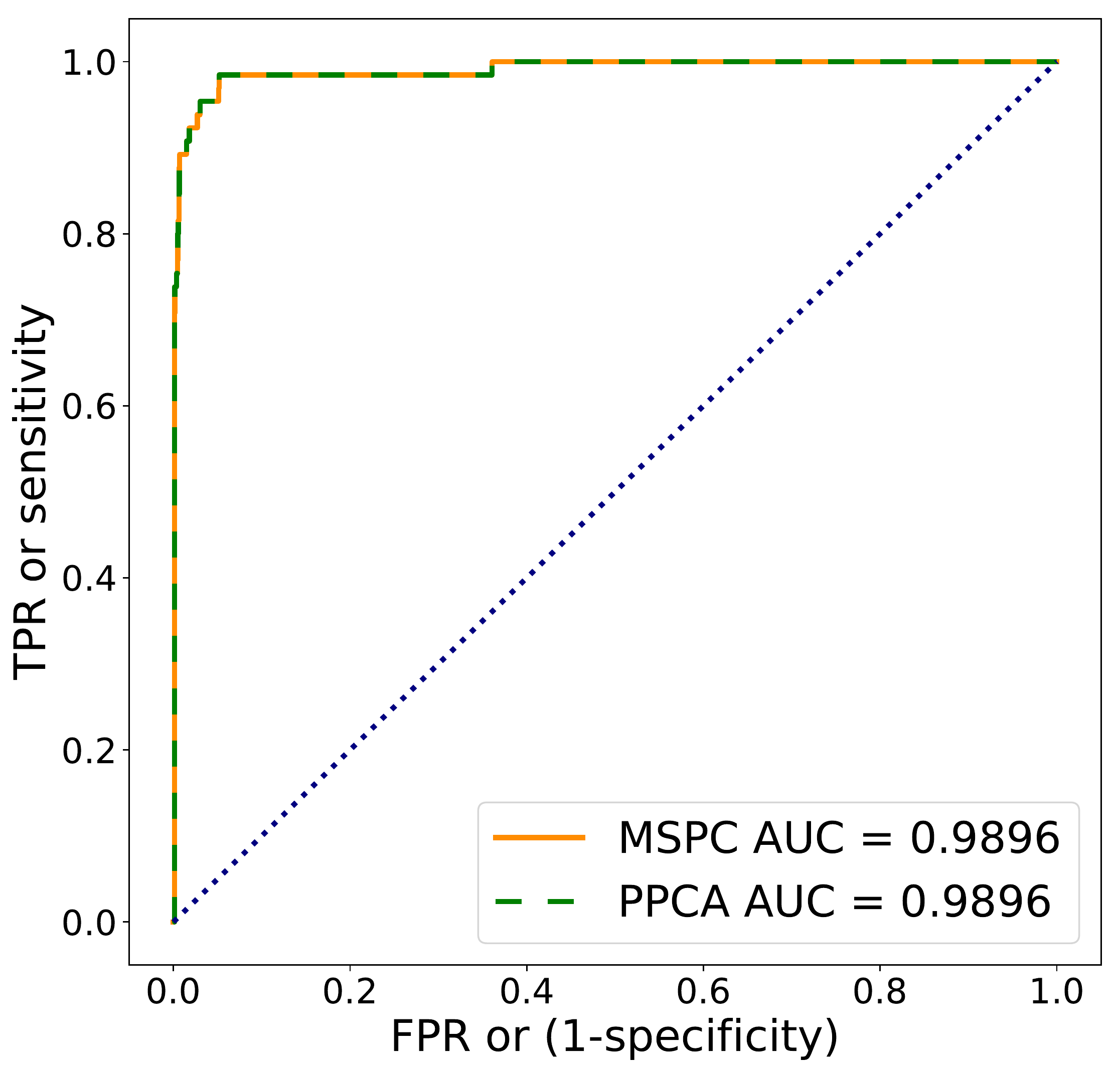}

    \\
a) DoS   & b) Scan44  \\

    
    \includegraphics[width=0.48\columnwidth]{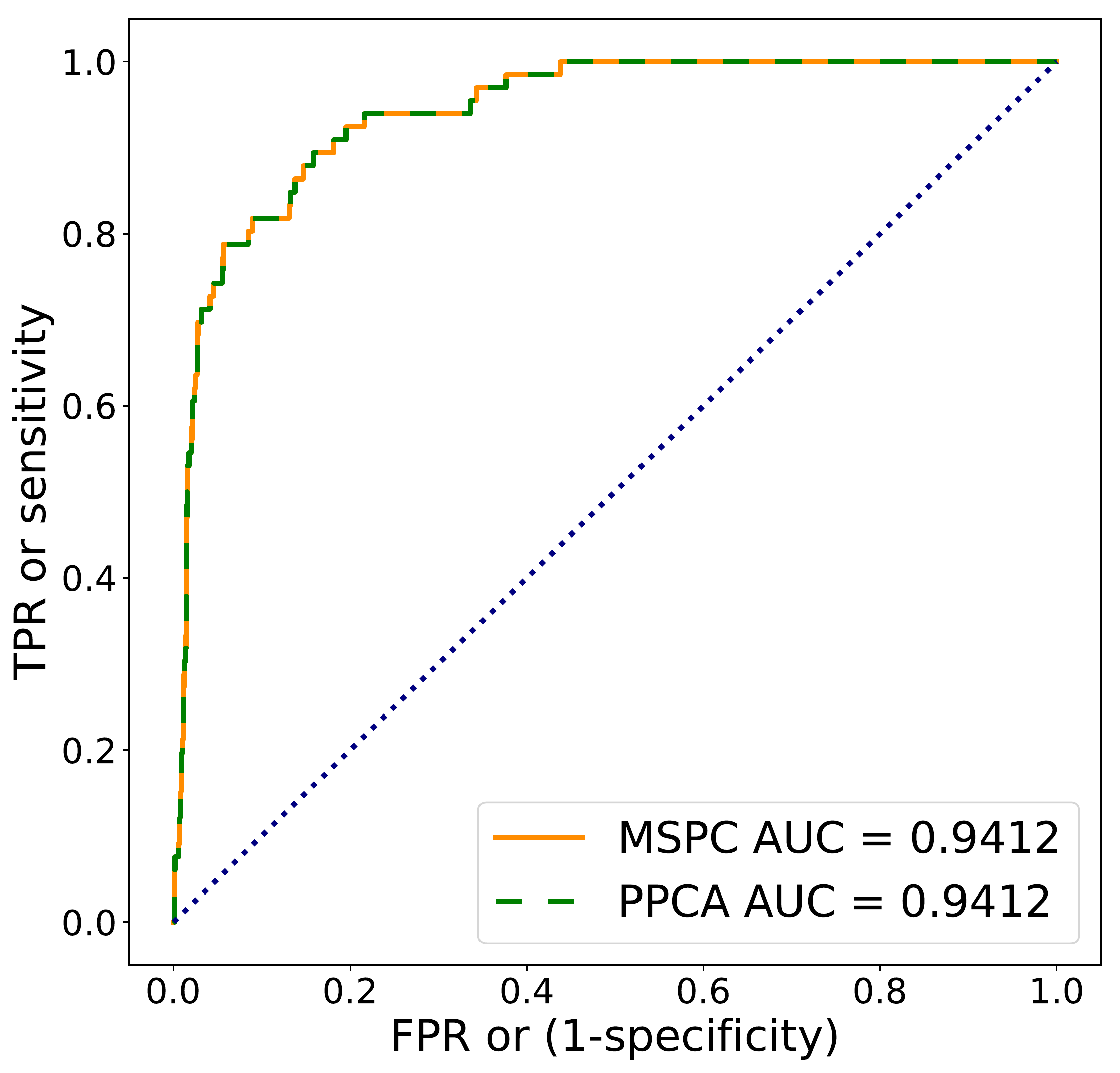}&
    \includegraphics[width=0.48\columnwidth]{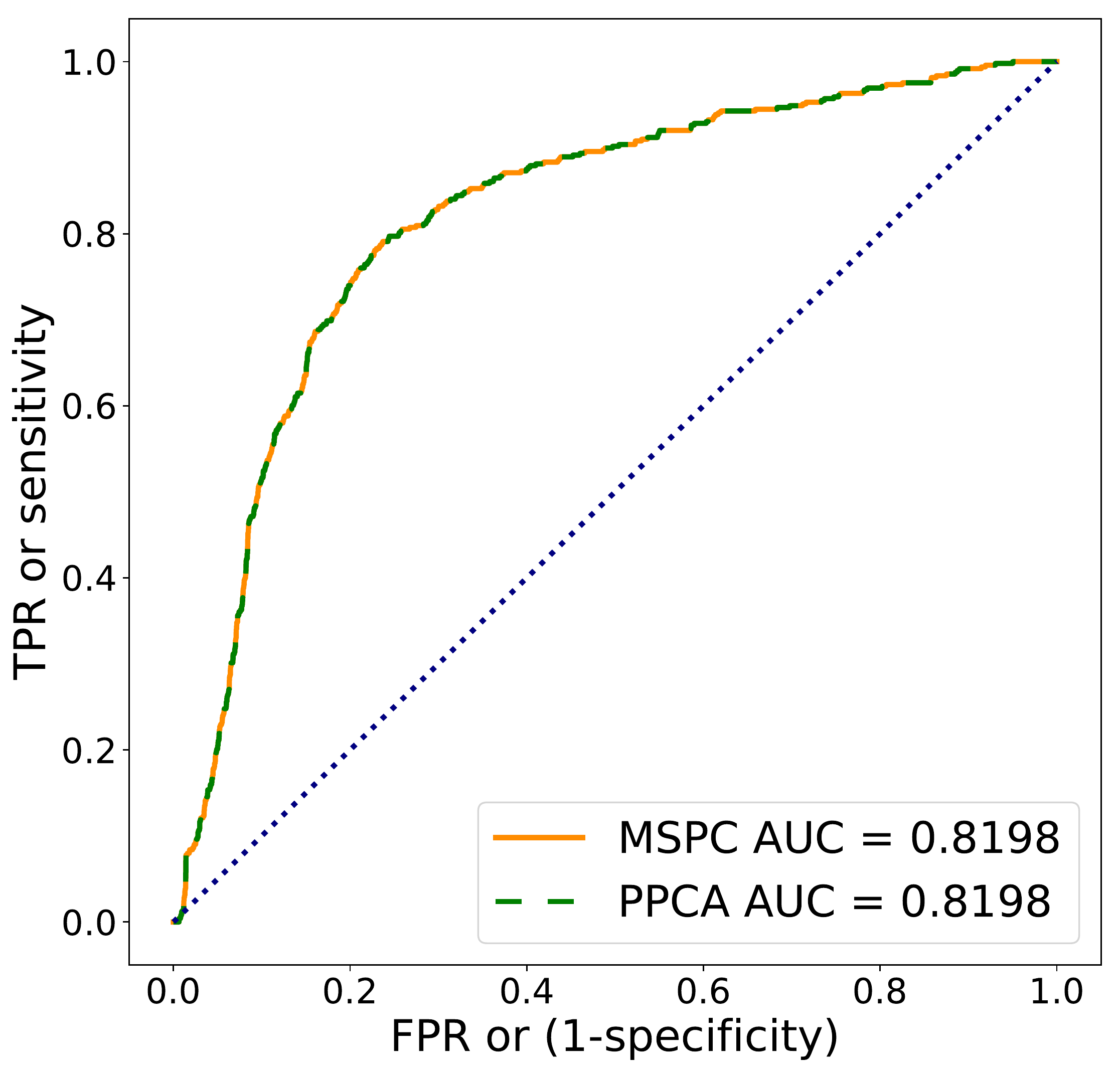}    
    \\
c) Scan11 &d) Neris Botnet  \\

    \end{tabular}
    \centering
    \caption{ROC curves for different attacks on UGR'16 using MSNM and PPCA with P=3. }
    \label{fig:AUC_UGR_MSNMvsPPCA}
\end{figure}

Figure~\ref{fig:Toy_MSPC_PPCA_comparison} shows the parallelism between MSNM and PPCA models in the synthetic test set. The same data points are identified as anomalies by both models, obtaining identical ROC curves and AUC values. The accuracy ($0.9858$) and the False Alarm Ratio ($0.012\%$) were also the same values for both models. The equivalence of both models is also tested on the UGR'16 dataset. Although the anomaly score is calculated for all testing data at once, we calculate the ROC curve and AUC for each attack type against the background traffic. Table~\ref{Table:UGR16_MSNMvsPPCA} includes identical results for both models when detecting different kind of attacks, using several numbers of principal components, $P=[1,2,...,5]$. For further details, Figure~\ref{fig:AUC_UGR_MSNMvsPPCA} shows the ROC curves obtained for both models in different attacks using $P=3$.

\begin{table*}[t!]
\caption{AUC Values for the attacks on UGR'16 using MSNM and PPCA}
\label{Table:UGR16_MSNMvsPPCA}
\centering
\begin{tabular}{lcccccccccc}
\toprule
&\multicolumn{2}{c}{1 P}&\multicolumn{2}{c}{2 P}&\multicolumn{2}{c}{3 P}&\multicolumn{2}{c}{4 P}&\multicolumn{2}{c}{5 P}\\
\cmidrule(lr){2-3}\cmidrule(lr){4-5}\cmidrule(lr){6-7}\cmidrule(lr){8-9}\cmidrule(lr){10-11}
Attack type& MSNM  &  PPCA  &  MSNM  &  PPCA  &  MSNM  &  PPCA &  MSNM  &  PPCA  &  MSNM  &  PPCA\\  
\midrule
DoS & 0.9118 &  0.9118 &  0.9089 &  0.9089 &  0.9089 &  0.9089 &  0.9097 &  0.9097 &  0.9091 &  0.9091 \\
Scan44 & 0.9903 &  0.9903 &  0.9902 &  0.9902 &  0.9896 &  0.9896 &  0.9882 &  0.9882 &  0.9880 &  0.9880 \\
Scan11 & 0.9384 &  0.9384 &  0.9390 &  0.9390 &  0.9412 &  0.9412 &  0.9318 &  0.9318 &  0.9303 &  0.9303 \\
Nerisbotnet & 0.8204 &  0.8204 &  0.8211 &  0.8211 &  0.8198 &  0.8198 &  0.8201 &  0.8201 &  0.8203 &  0.8203 \\
UDPscan & 0.7826 &  0.7826 &  0.7844 &  0.7844 &  0.7707 &  0.7707 &  0.7707 &  0.7707 &  0.7727 &  0.7727 \\
SSHscan & 0.5593 &  0.5593 &  0.5624 &  0.5624 &  0.5569 &  0.5569 &  0.5588 &  0.5588 &  0.5614 &  0.5614 \\
Spam & 0.4669 &  0.4669 &  0.4610 &  0.4610 &  0.4588 &  0.4588 &  0.4585 &  0.4585 &  0.4512 &  0.4512 \\
\bottomrule
\end{tabular}
\end{table*}




    



\subsubsection{Using the regularization and reconstruction error terms to detect different types of anomalies}
\begin{figure*}[t]
    \centering
     \setlength{\tabcolsep}{1.7pt} 
\setlength\fboxsep{0pt}
\pdfpxdimen=\dimexpr 1 in/72\relax

    \begin{tabular}{ccc}
    \includegraphics[width=0.33\textwidth]{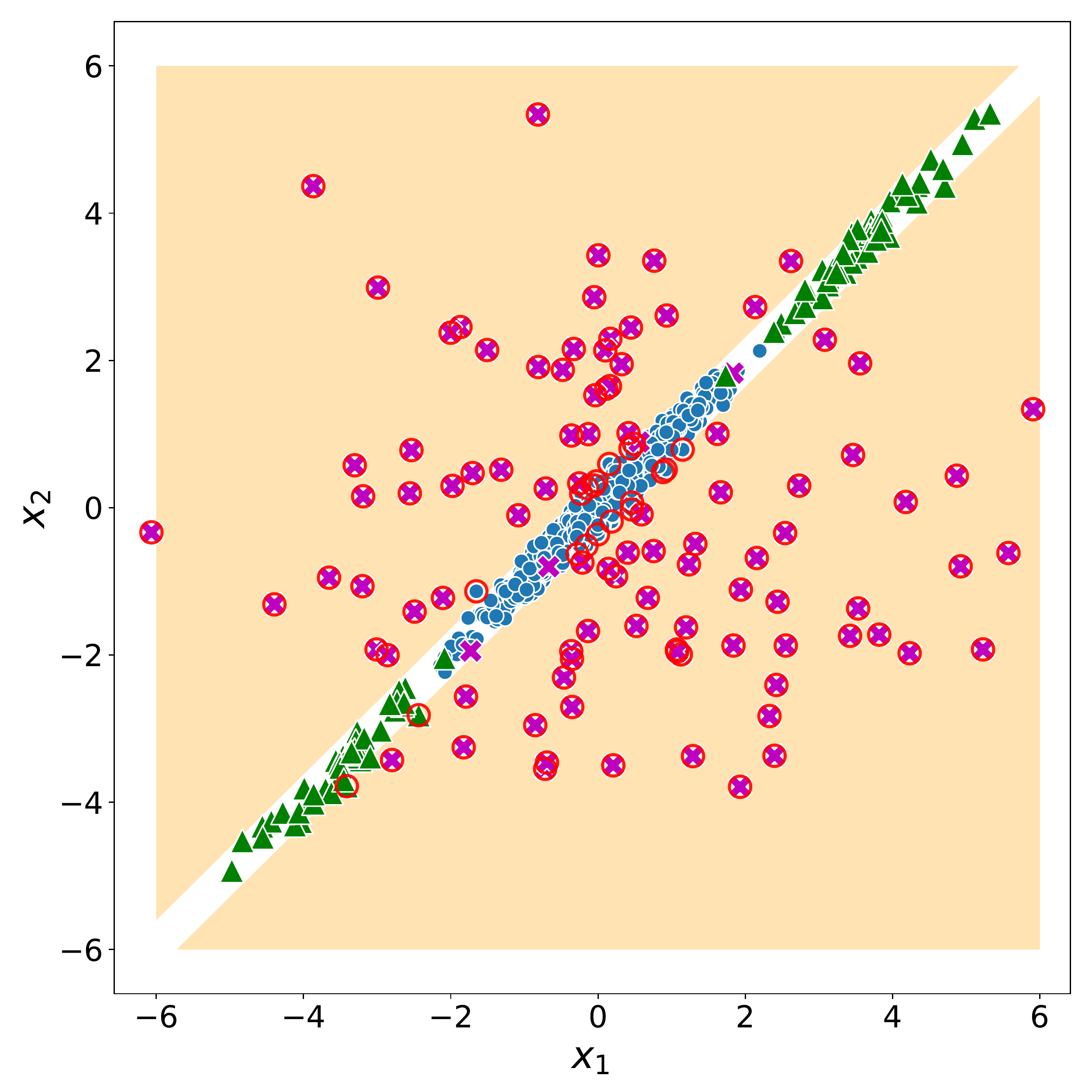}
    &
    \includegraphics[width=0.33\textwidth]{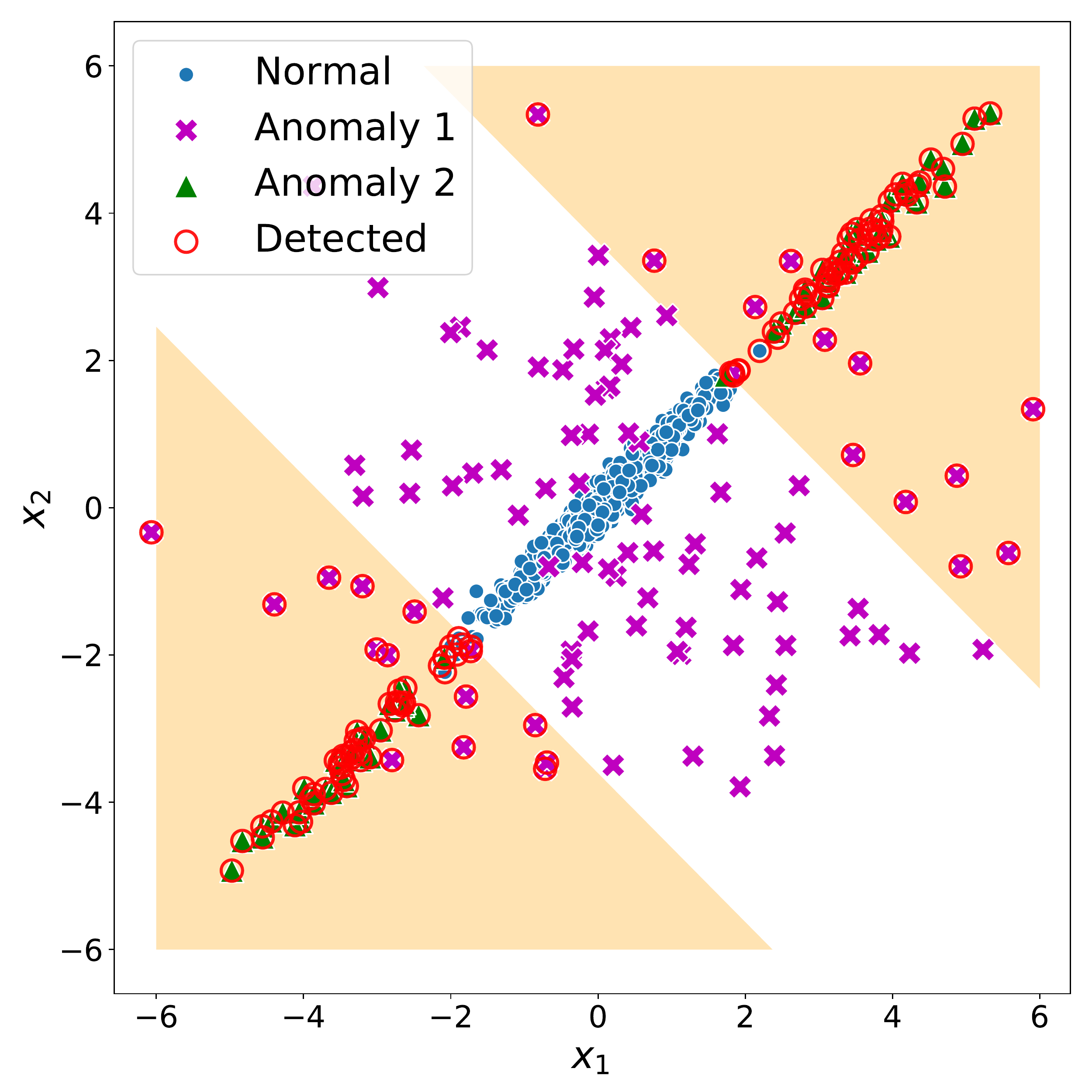}
    &
    \includegraphics[width=0.33\textwidth]{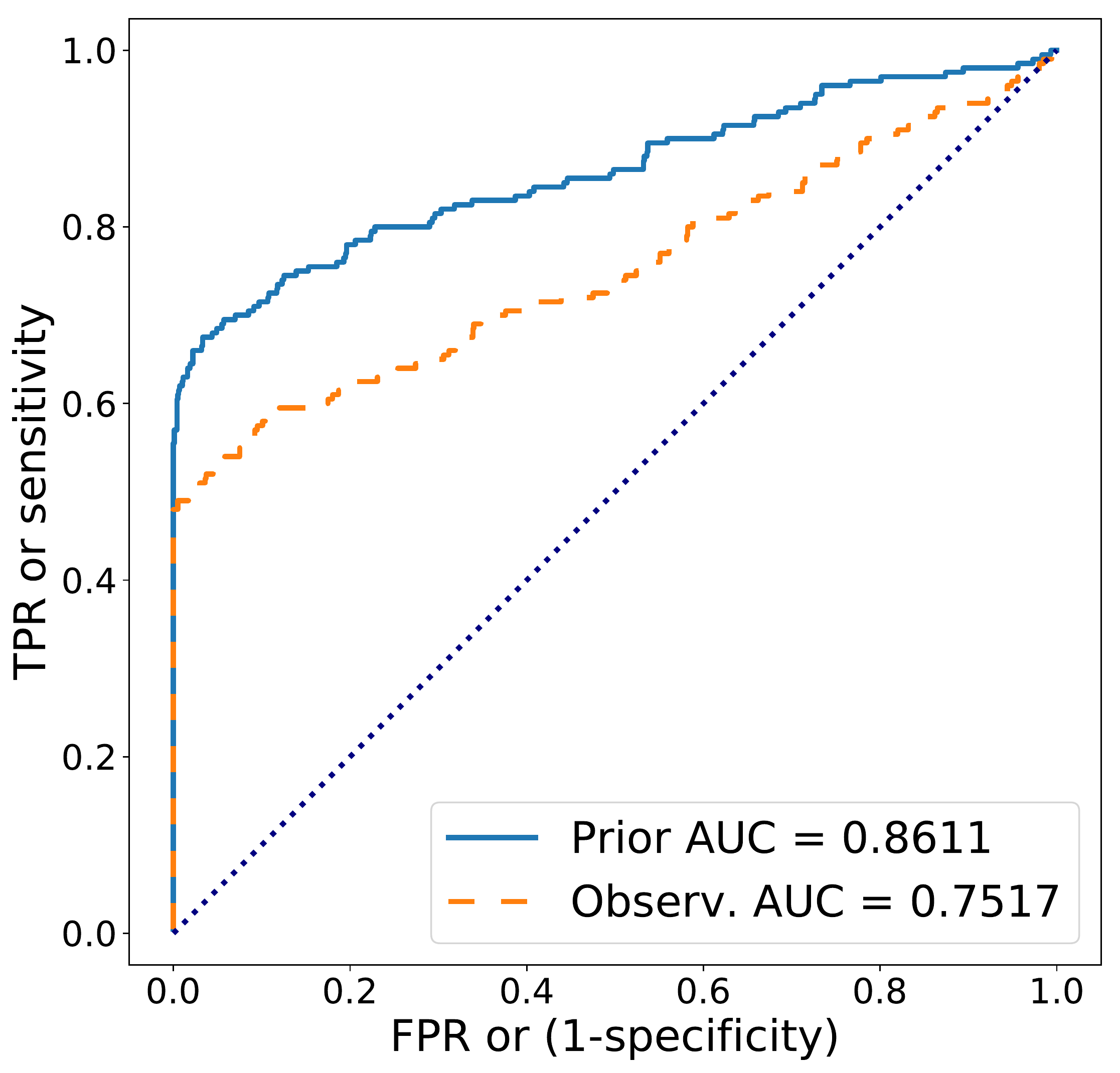}
    
    \\

a) Observation model only   & b) Prior model only & c) ROC curves for both terms 
    
    \end{tabular}
    \centering
    \caption{Detected anomalies in the synthetic dataset using only a single term. Blue circles identify normal data, fuchsia \textbf{x}s and green triangles identify different types of anomalies. Red circumferences mark the points detected as anomalies. The orange shadows cover the detection area in each case.} 
    \label{fig:Toy_KLvsError}
\end{figure*}

As previously indicated, both terms in Eq.~\eqref{eq:PPCA_MSPC_relation} are able to identify different behaviour in the data. To quickly gain an overview of how both terms work, Figure~\ref{fig:Toy_KLvsError} depicts the anomalies in the synthetic dataset that are detected using a single PPCA/MSNM term only. Note that when using a single term, the threshold is the $99^{th}$-percentile of the calibration set for that term. When we only use the reconstruction error term ---Eq.~\eqref{eq:error}---, we can see in Figure~\ref{fig:Toy_KLvsError}(a) how we are unable to identify anomalies that have been generated following the linear model. Note that this case is equivalent to using a simple PCA analysis. Using the regularization term only ---Eq.~\eqref{eq:regularizationterm}---, see Figure~\ref{fig:Toy_KLvsError}(b), results in a similar problem. This term is calculated using only the latent $\hat{z}(\delta)$, therefore all data points whose latent value is within the distribution of the normal data remains undetected. To effectively detect all anomalies, it is clearly necessary to use both terms.


When it comes to complex networking data, such as those in UGR'16, it is difficult to predict which anomalies will be correctly detected by the two terms in the PPCA/MSNM model. Table~\ref{Table:UGR16_KLvsError} shows that different attacks are captured differently by both terms. Some attacks (DoS, Scan44) are captured correctly by both terms. Neris Botnet traffic can be identified using the reconstruction error term, but cannot usually be identified with the regularization term. 

The results for Scan11 and UDP Scan attacks show that the ability of the model to identify anomalies using the error reconstruction or regularization terms depends on the number of principal components used. We have included Fig~\ref{fig:Scan11_KLvsError} as an example of where the discriminative power of the regularization term increases with $P$, even outperforming the error reconstruction term performance. Note that the error reconstruction term is not severely affected by the number of $P$ because the explained variance of the model is low, and therefore the error reconstruction for most anomalies remains high. As the number of $P$ increases, the latent space is able to capture more features of the data, also allowing anomalies in the latent domain to better identified.

\begin{table*}

\caption{AUC Values for the attacks on UGR'16 using divergence (Prior) and error (Observ.) terms separately.}

\centering
\setlength\tabcolsep{3pt}
\begin{tabular}{lcccccccccc}
\toprule
&\multicolumn{2}{c}{1 P}&\multicolumn{2}{c}{2 P}&\multicolumn{2}{c}{3 P}&\multicolumn{2}{c}{4 P}&\multicolumn{2}{c}{5 P}\\
\cmidrule(lr){2-3}\cmidrule(lr){4-5}\cmidrule(lr){6-7}\cmidrule(lr){8-9}\cmidrule(lr){10-11}
Attack type &   Prior &  Observ.&   Prior &  Observ. &   Prior & Observ. &   Prior &  Observ. &  Prior &  Observ. \\
\midrule
DoS             & 0.5242 &   \bf0.9118 & \bf0.9191 &   0.9085 & \bf0.9178 &   0.9085 & 0.9050 &   \bf0.9095 & \bf0.9152 &   0.9087 \\
Scan44          & 0.8747 &   \bf0.9903 & 0.8483 &   \bf0.9902 & \bf0.9968 &   0.9895 & \bf0.9960 &   0.9878 & \bf0.9957 &   0.9875 \\
Scan11          & 0.4966 &   \bf0.9385 & 0.3771 &   \bf0.9391 & 0.7714 &   \bf0.9416 & \bf0.9691 &   0.9304 & \bf0.9691 &   0.9285 \\
Nerisbotnet     & 0.4358 &   \bf0.8207 & 0.3686 &   \bf0.8214 & 0.4741 &   \bf0.8199 & 0.4480 &   \bf0.8203 & 0.4926 &   \bf0.8204 \\
UDPscan & 0.6618 &   \bf0.7817 & 0.7084 &   \bf0.7838 & \bf0.8886 &   0.7682 & \bf0.8659 &   0.7674 & \bf0.8372 &   0.7698 \\
\midrule
Mean & 0.5986 &   \bf0.8886 & 0.6443 &   \bf0.8886 & 0.8097 &   \bf0.8855 & 0.8368 &  \bf 0.8831 & 0.8419 &  \bf 0.8830 \\
\bottomrule
\end{tabular}
\label{Table:UGR16_KLvsError}
\end{table*}

\begin{table*}[]
    \centering

    \caption{Accuracy Values for the attacks on UGR'16 using divergence (Prior) and error (Observ.) terms separately.}
    \setlength\tabcolsep{3pt}
\begin{tabular}{lcccccccccc}
\toprule
&\multicolumn{2}{c}{1 P}&\multicolumn{2}{c}{2 P}&\multicolumn{2}{c}{3 P}&\multicolumn{2}{c}{4 P}&\multicolumn{2}{c}{5 P}\\
\cmidrule(lr){2-3}\cmidrule(lr){4-5}\cmidrule(lr){6-7}\cmidrule(lr){8-9}\cmidrule(lr){10-11}
Attack type &   Prior &  Observ.&   Prior &  Observ. &   Prior & Observ. &   Prior &  Observ. &  Prior &  Observ. \\
\midrule
Dos             & \bf0.9237 &   0.8946 & \bf0.9655 &   0.8939 & \bf0.9273 &   0.8932 & 0.8939 &   \bf0.8943 & \bf0.9287 &   0.8922 \\
scan44          & \bf0.9507 &   0.9011 & \bf0.9713 &   0.9008 & \bf0.9320 &   0.9004 & 0.8979 &  \bf 0.9015 & \bf0.9347 &   0.8991 \\
scan11          &\bf 0.9460 &   0.8999 & \bf0.9667 &   0.8995 & \bf0.9300 &   0.8992 & 0.8974 &   \bf0.9001 & \bf0.9337 &   0.8977 \\
nerisbotnet     & \bf0.9022 &   0.8801 & \bf0.9223 &   0.8798 & 0.8850 &   \bf0.8795 & 0.8541 &   \bf0.8806 & \bf0.8880 &   0.8788 \\
anomaly-udpscan & \bf0.9520 &   0.9001 & \bf0.9728 &   0.8998 & \bf0.9312 &   0.8995 & 0.8969 &   \bf0.9005 & \bf0.9340 &   0.8981 \\
\midrule
mean            & \bf0.9349 &   0.8952 & \bf0.9597 &   0.8948 & \bf0.9211 &   0.8943 & 0.8880 &   \bf0.8954 & \bf0.9238 &   0.8932 \\
\bottomrule
\end{tabular}
    \label{tab:UGR16_KLvsError_ACC}
\end{table*}

\begin{table*}[]
    \centering

    \caption{Mean false alarm ratio on UGR'16 using divergence (Prior) and error (Observ.) terms separately.}
    \setlength\tabcolsep{3pt}
\begin{tabular}{cccccccccc}
\toprule
\multicolumn{2}{c}{1 P}&\multicolumn{2}{c}{2 P}&\multicolumn{2}{c}{3 P}&\multicolumn{2}{c}{4 P}&\multicolumn{2}{c}{5 P}\\
\cmidrule(lr){1-2}\cmidrule(lr){3-4}\cmidrule(lr){5-6}\cmidrule(lr){7-8}\cmidrule(lr){9-10}
  Prior &  Observ.&   Prior &  Observ. &   Prior & Observ. &   Prior &  Observ. &  Prior &  Observ. \\
 \bf0.0474 &   0.0995 &\bf 0.0267 &   0.0998 & \bf0.0684 &   0.1002 & 0.1027 &   \bf0.0992 & \bf0.0656 &   0.1016 \\
\bottomrule
\end{tabular}
    \label{tab:UGR16_KLvsError_f1}
\end{table*}



    
    


     


\begin{figure*}[t]
    \centering
     \setlength{\tabcolsep}{1.7pt} 
\setlength\fboxsep{0pt}
\pdfpxdimen=\dimexpr 1 in/72\relax

    \begin{tabular}{cc}
    \includegraphics[width=0.31\textwidth]{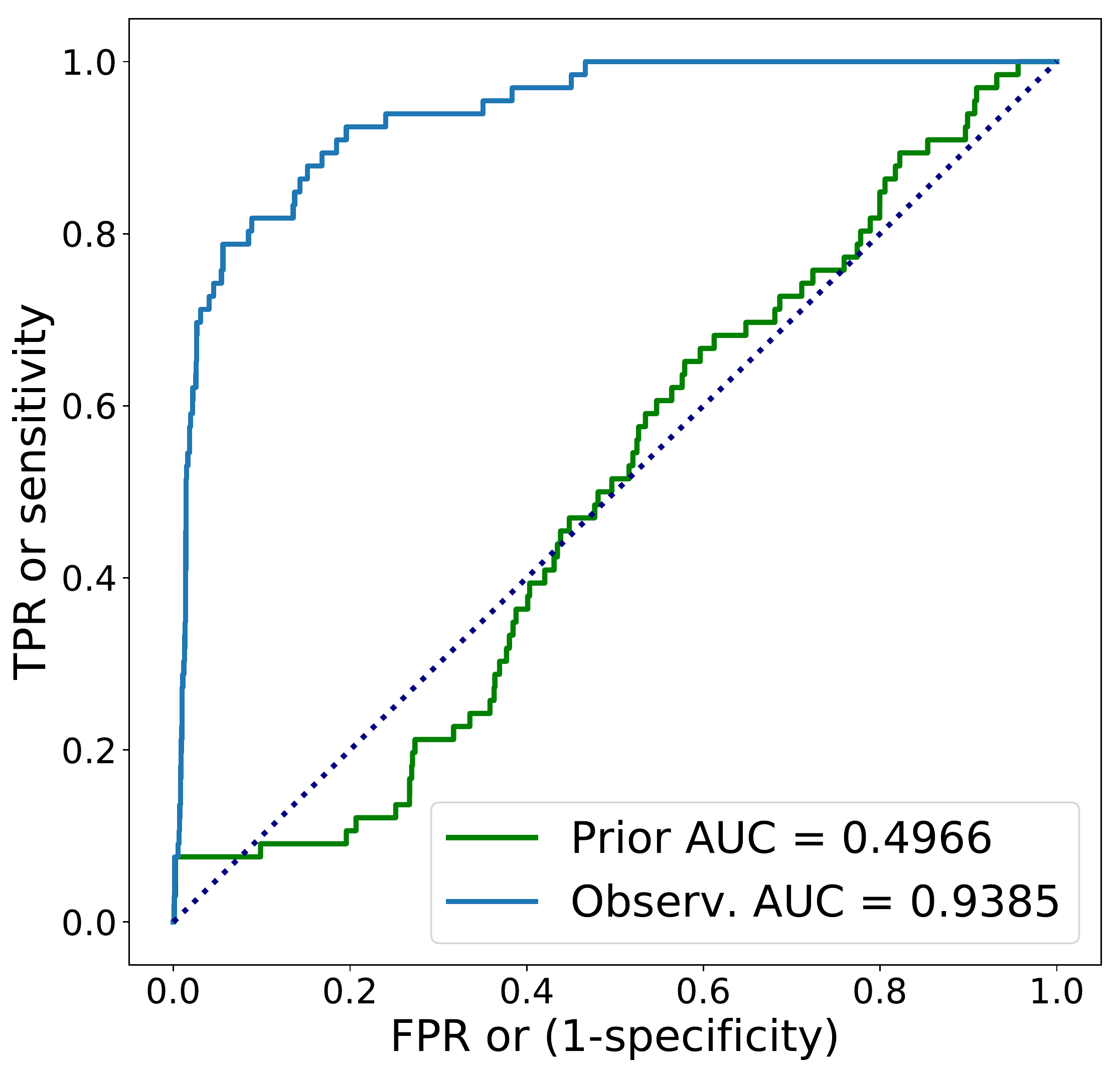}
    &
    \includegraphics[width=0.31\textwidth]{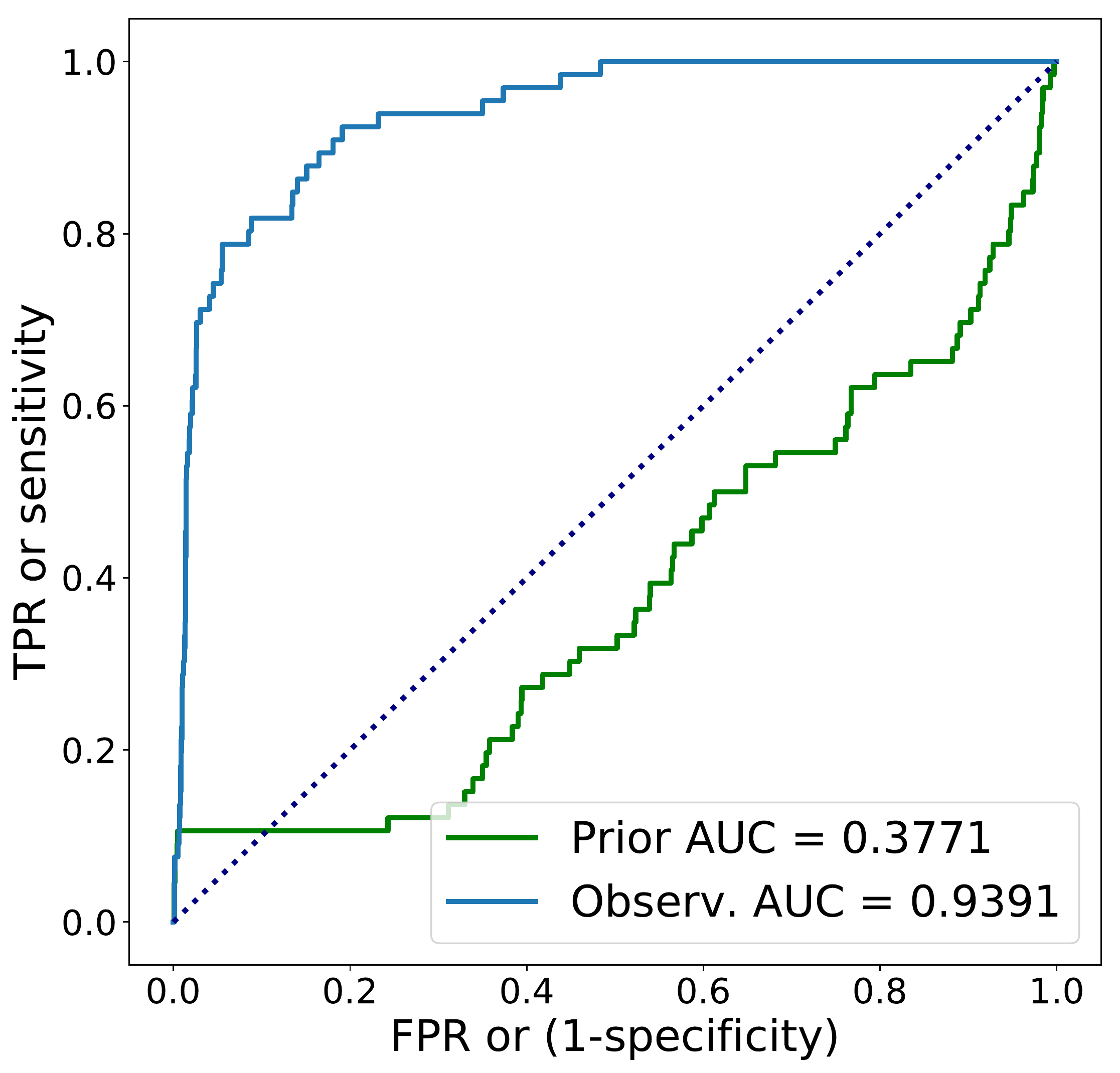}

    \\
     
a) 1 P   & b) 2 P 

    \end{tabular}
    
        \begin{tabular}{ccc}
    \includegraphics[width=0.31\textwidth]{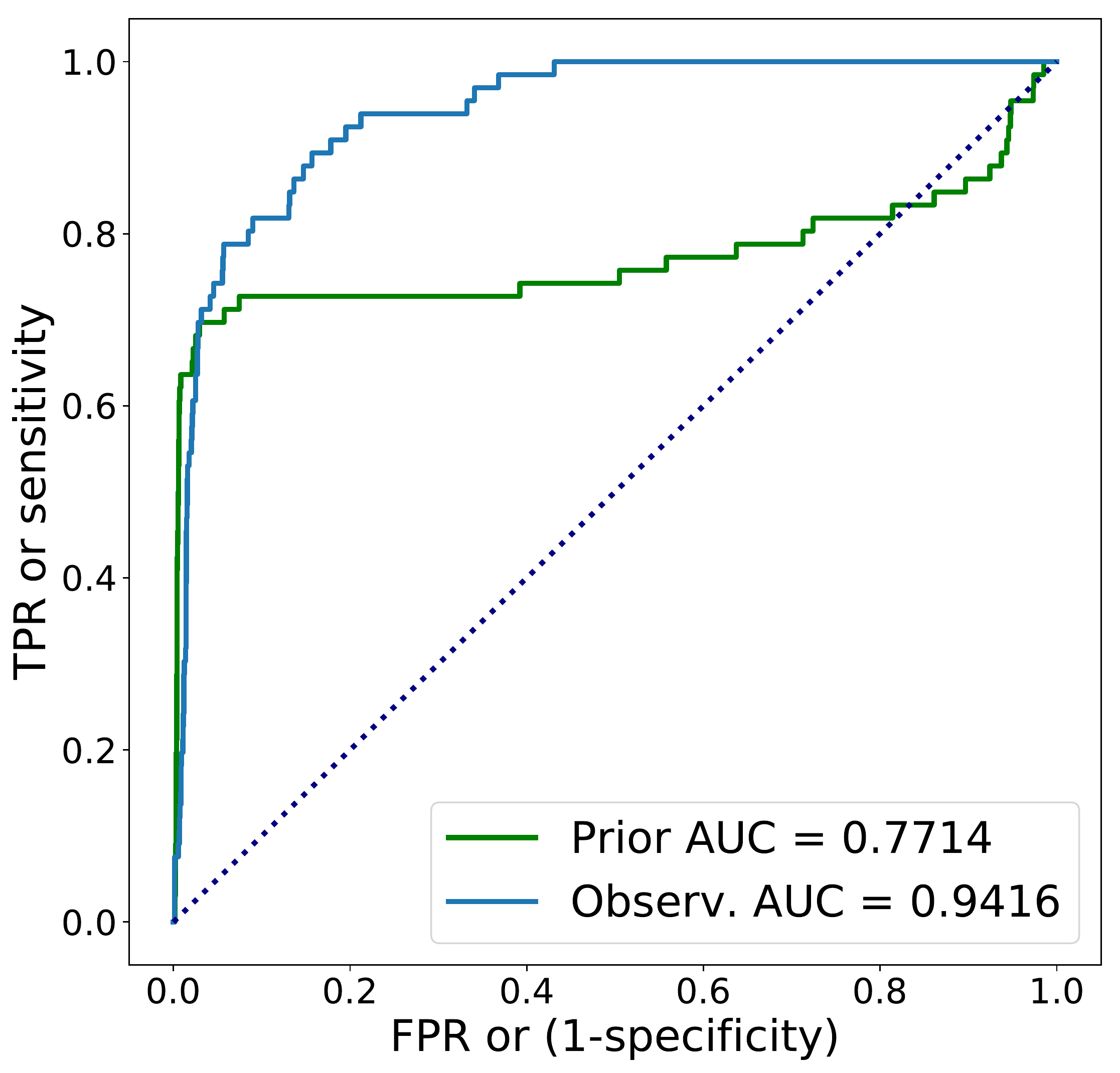}
&
    
    \includegraphics[width=0.31\textwidth]{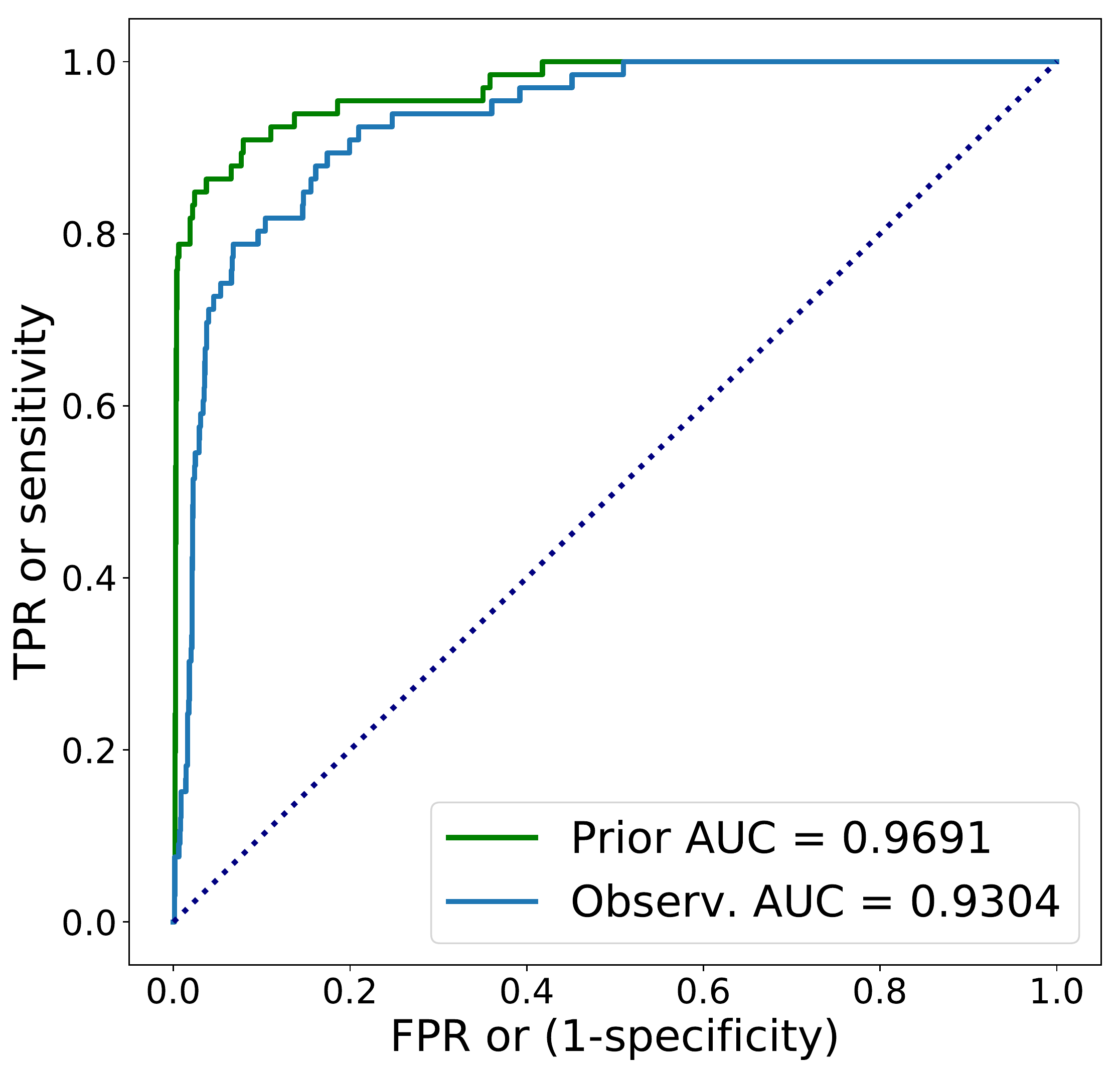}
    &
    \includegraphics[width=0.31\textwidth]{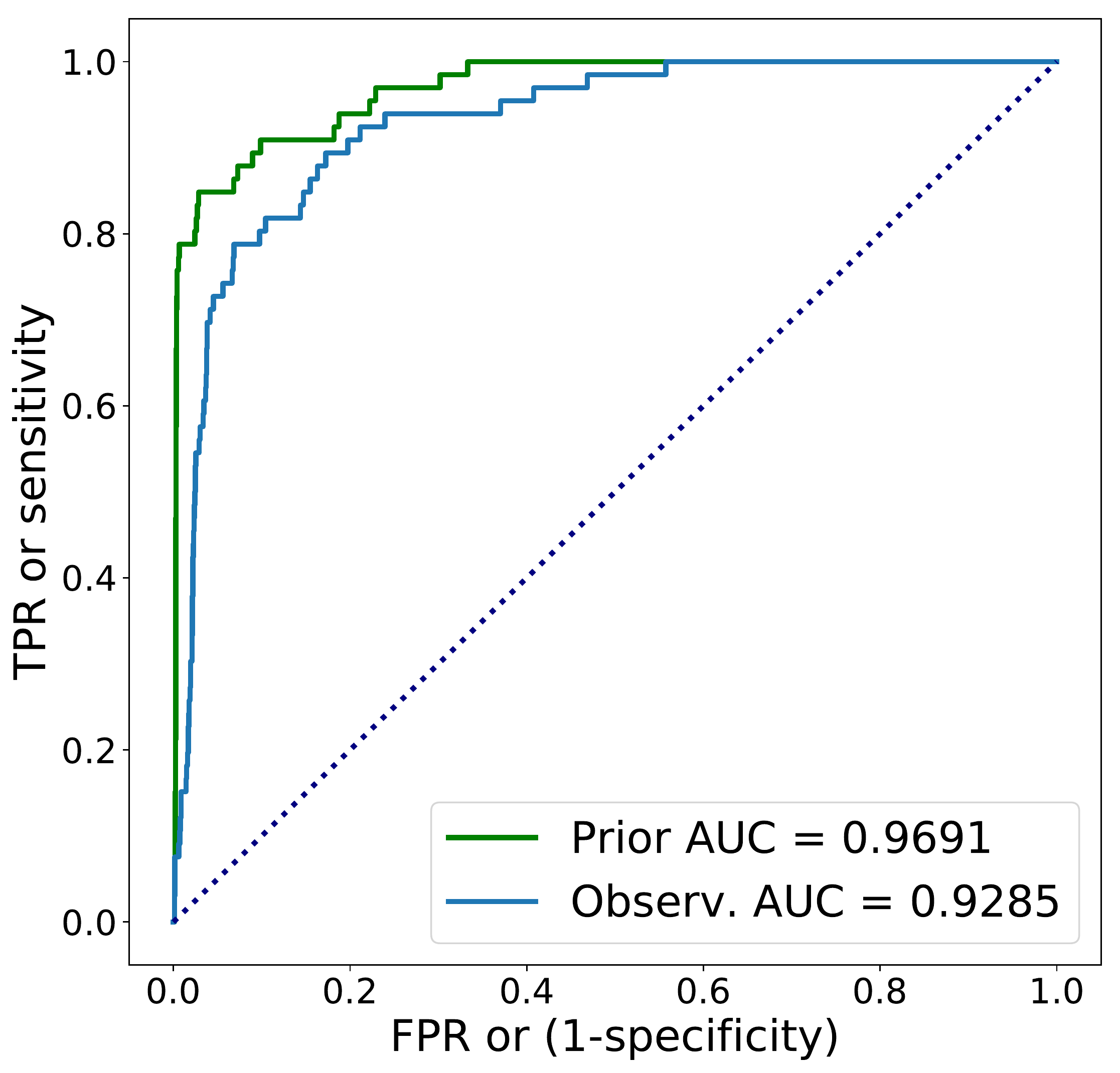}

    \\
     
 c) 3 P  & d) 4 P & e) 5 P

    \end{tabular}
    \centering
    \caption{Evolution of the ROC curves of both divergence and error terms on the Scan11 attack for different latent space sizes.}
    \label{fig:Scan11_KLvsError}
\end{figure*}

\subsubsection{Weighting reconstruction error and regularization terms}

The previous experiments have shown that both terms in Eq.~\eqref{eq:PPCA_MSPC_relation} are relevant and should be used. Furthermore, a single term is not usually able to correctly identify all types of anomalies. The forthcoming question is how to combine them to obtain a single anomaly score. 

On the one hand, the work in \cite{camacho2017traffic} suggested the use of the significance levels $U\!C\!L_Q$ and $U\!C\!L_D$ and a weighting parameter calculated as $P/M$ (see Section~\ref{sec:MSNM}). On the other hand, PPCA provides a probabilistic interpretation for the weighting parameter, combining both terms according to their contribution to the marginal likelihood ---see Eq.~\eqref{eq:split_marginal}---. In this section, we compare the MSNM approach in Eq.~\eqref{MSNM_score} with the PPCA approach in Eq.~\eqref{eq:split_marginal}.


\begin{table*}
\caption{AUC Values for the attacks on UGR'16 using different number of PCs and models: MSNM ---Eq.~\eqref{MSNM_score}--- and PPCA ---Eq.~\eqref{eq:split_marginal}---.}
\label{table:AUC_UCLvsPPCA}
\centering
\begin{tabular}{lcccccccccc}
\toprule
&\multicolumn{2}{c}{1 P}&\multicolumn{2}{c}{2 P}&\multicolumn{2}{c}{3 P}&\multicolumn{2}{c}{4 P}&\multicolumn{2}{c}{5 P}\\
\cmidrule(lr){2-3}\cmidrule(lr){4-5}\cmidrule(lr){6-7}\cmidrule(lr){8-9}\cmidrule(lr){10-11}
Attack type &  MSNM  &   PPCA &   MSNM &   PPCA &   MSNM &   PPCA &   MSNM &   PPCA &   MSNM &   PPCA \\
\midrule
DoS             & 0.9118 & 0.9118 & 0.9085 & \bf0.9089 & 0.9085 & \bf0.9089 & 0.9095 & \bf0.9097 & 0.9087 & \bf0.9091 \\
Scan44          & 0.9903 & 0.9903 & 0.9902 & 0.9902 & 0.9895 & \bf0.9896 & 0.9878 & \bf0.9882 & 0.9875 & \bf0.9880 \\
Scan11          & \bf0.9385 & 0.9384 & \bf0.9391 & 0.9390 & \bf0.9415 & 0.9412 & 0.9305 & \bf0.9318 & 0.9286 & \bf0.9303 \\
Nerisbotnet     & \bf0.8207 & 0.8204 & \bf0.8214 & 0.8211 & \bf0.8199 & 0.8198 & \bf0.8203 & 0.8201 & \bf0.8204 & 0.8203 \\
UDPscan & 0.7817 & \bf0.7826 & \bf0.7838 & \bf0.7844 & 0.7683 & \bf0.7707 & 0.7674 & \bf0.7707 & 0.7700 & \bf0.7727 \\
\midrule
Mean            & 0.8886 & \bf0.8887 & 0.8886 & \bf0.8887 & 0.8855 & \bf0.8860 & 0.8831 & \bf0.8841 & 0.8830 & \bf0.8841 \\
\bottomrule
\end{tabular}
\end{table*}

\begin{table*}[]
    \centering

    \caption{Accuracy Values for the attacks on UGR'16 using different number of PCs and models: MSNM ---Eq.~\eqref{MSNM_score}--- and PPCA ---Eq.~\eqref{eq:split_marginal}---.}
\begin{tabular}{lcccccccccc}
\toprule
{} &  MSNM  &   PPCA &   MSNM &   PPCA &   MSNM &   PPCA &   MSNM &   PPCA &   MSNM &   PPCA \\
\midrule
Dos             & 0.8946 & \bf0.8947 & 0.8939 & \bf0.8946 & 0.8932 & \bf0.9655 & 0.8944 & \bf0.8947 & 0.8922 & \bf0.8932 \\
scan44          & 0.9011 & \bf0.9012 & 0.9008 & \bf0.9011 & 0.9004 & \bf0.9713 & 0.9015 & \bf0.9017 & 0.8991 & \bf0.9004 \\
scan11          & 0.8999 & \bf0.9000 & 0.8995 & \bf0.8999 & 0.8992 & \bf0.9667 & 0.9001 & \bf0.9003 & 0.8977 & \bf0.8992 \\
nerisbotnet     & 0.8801 & \bf0.8802 & 0.8798 & \bf0.8801 & 0.8794 & \bf0.9223 & 0.8806 & \bf0.8807 & 0.8788 & \bf0.8795 \\
anomaly-udpscan & 0.9001 & \bf0.9003 & 0.8998 & \bf0.9001 & 0.8995 & \bf0.9728 & 0.9005 & \bf0.9007 & 0.8981 & \bf0.8995 \\
\midrule
mean            & 0.8952 & \bf0.8953 & 0.8948 & \bf0.8952 & 0.8943 & \bf0.9597 & 0.8954 & \bf0.8956 & 0.8932 & \bf0.8943 \\
\bottomrule
\end{tabular}
    \label{tab:AUC_UCLvsPPCA_ACC}
\end{table*}

\begin{table*}[]
    \centering

    \caption{Mean false alarm ratio on UGR'16 using different number of PCs and models: MSNM ---Eq.~\eqref{MSNM_score}--- and PPCA ---Eq.~\eqref{eq:split_marginal}---.}
    \setlength\tabcolsep{3pt}
\begin{tabular}{cccccccccc}
\toprule
\multicolumn{2}{c}{1 P}&\multicolumn{2}{c}{2 P}&\multicolumn{2}{c}{3 P}&\multicolumn{2}{c}{4 P}&\multicolumn{2}{c}{5 P}\\
\cmidrule(lr){1-2}\cmidrule(lr){3-4}\cmidrule(lr){5-6}\cmidrule(lr){7-8}\cmidrule(lr){9-10}
 MSNM  &   PPCA &   MSNM &   PPCA &   MSNM &   PPCA &   MSNM &   PPCA &   MSNM &   PPCA \\
 0.0995 & \bf0.0994 & \bf0.0998 &   0.1000 & 0.1002 &   \bf0.1000 & 0.0992 & \bf0.0989 & \bf0.1016 & 0.1018 \\
\bottomrule
\end{tabular}
    \label{tab:AUC_UCLvsPPCA_f1}
\end{table*}


In the synthetic dataset, both strategies obtain a similar result in AUC: 0.9974 for PPCA and 0.9973 for MSNM. Note that in both cases, the combination of both terms obtains a better AUC than the values reported in the previous section for each term separately. 

The difference between the results from PPCA and MSNM is more clearly shown in Table \ref{table:AUC_UCLvsPPCA}, where the UGR'16 dataset is analyzed. Even when the MSNM approach results in slightly better AUC values for some combinations of attacks and number of P, the PPCA AUC score yields a better mean value in each case. While the differences are subtle, they indicate that PPCA is better when it comes to combining information from both regularization and reconstruction error terms. Looking at Table~\ref{table:AUC_UCLvsPPCA}, we can observe that PPCA obtains better values than MSNM in those attacks where the regularization term is more informative. In addition, note that while a single term might be better at recognising a given attack type, the mean AUC values are always higher than those obtained using only a single term.


\section{Conclusions}
\label{sec:conclusions}

In this paper we have analyzed the use of the generative model known as probabilistic PCA (PPCA) for detection of anomalies in network security. Specifically, we have provided a detailed mathematical model that connects MSNM, a well-known framework for network anomaly detection, and PPCA. The generative PPCA model provides a probabilistic point of view to explain the MSNM framework and the meaning of its principal elements: The use of $Q$ and $D$ statistics, which are derived as an error reconstruction term and a regularization term, respectively, in the PPCA formulation. 

Understanding the role of both terms in the anomaly detection process is a key step towards the correct use of generative models in this security research field. Specifically, a direct application is that of correctly using other generative models like VAEs and GANs. In a review of research works that use these models, we note that while the error reconstruction term is widely used due to its high anomaly detection capability, the regularization term is often forgotten or discarded. We have theoretically and experimentally assessed that both terms are relevant and capture complementary information. This implies that they should be used together for a robust anomaly detection. 

In addition, the PPCA generative framework provides a combination of both terms that considers their contribution to the marginal distribution $\p(\bx)$ with a probabilistic interpretation, thus offering a non-empirical solution for the weighting parameter required by MSNM.

Although the linear PPCA generative model is easy to understand and helps obtain the above conclusions, its simplicity limits its generalization to non-linear data. Non-linearity is often present in real traffic data and would be difficult to capture and detect.
Also, PPCA inherits some of the disadvantages from PCA, which is quite common when working with latent space models. Even with a linear model, the latent combination of the original features is not easy to interpret. So, as shown in Figure~\ref{fig:Scan11_KLvsError}, the choice of the latent space size $P$ is critical for the detection of some attacks. 
While the use of linear detection models is useful in the later diagnosis of network incidents, this paper intends to establish a first step for more complex generative approaches to network anomaly detection. Further research on the combination of both error reconstruction and regularization terms for those models is also needed. Finally, the choice of the threshold is a well-known problem for anomaly detection. The calibration set provides a benchmark to choose the decision boundary, but it should be determined according to the confidence in the calibration data. The Gaussian form of $p(x|\bW_{\mbox{\small ML}},\sigma^2_{\mbox{\small ML}})$ allows us to choose the confidence-based threshold. Although the threshold might also be experimentally determined with the use of testing sets, this approach relies on known attacks, and therefore does not provide information about the optimal threshold for new or unknown attacks. Finally, system requirements usually determine the threshold to use in industry applications. A high cost of undetected false negatives might induce the use of a lower threshold that will produce a higher number of false positives. When combining both decision terms, the use of a combined or separated threshold for each term is also an open field for future research. 


\section*{Acknowledgment}
This work was sponsored in part by the Agencia Estatal de Investigación under project PID2019-105142RB-C22/AEI/10.13039/501100011033,  Spanish MINECO (Ministerio
de Economía y Competitividad) project 
TIN2017-83494-R and FEDER/Junta de Andalucia-Consejería de Transformación Económica, Industria y Universidades/ project A-TIC-215-UGR18.
The work by Fernando P\'{e}rez-Bueno was sponsored by Ministerio de Econom\'{i}a, Industria y Competitividad under FPI contract BES-2017-081584.

The authors would like to thank Daniel Cort\'{e}s Troya for his collaboration in the early stages of this work.

\appendix
\section{}
\subsection{Laplace approximation\label{app:laplace}}

It follows that:
\begin{align}
     \p(\bx)&=\int\p(\bz)\p(\bx|\bz)\mbox{d}\bz=\int\exp[\ln\p(\bz)\p(\bx|\bz)]\mbox{d}\bz\label{eq:ap:marginal}
\end{align}
where $f(\bz)=\ln(\p(\bz)\p(\bx|\bz))$ is a quadratic function that can be expanded around the maximum a posteriori (MAP)
\begin{equation}
    \hat \bz= \arg\max_\bz f(\bz) =\arg\max_\bz \ln(\p(\bz)\p(\bx|\bz))
\end{equation}
to obtain
\begin{align}
 \p(\bz)\p(\bx|\bz)&=\exp[\ln\p(\bz)\p(\bx|\bz)]\\
 &\propto\exp[f(\hat \bz)-\frac{1}{2}(\bz-\hat\bz)^\T(\bI+\frac{1}{\sigma^2}\bW^\T\bW)(\bz-\hat\bz)], \nonumber
\end{align}
which produces, from Eq. (\ref{eq:ap:marginal}), $\ln \p(\bx)=f(\hat\bz) + \mbox{const}$, and so
\begin{equation}
    \ln \p(\bx|\bW,\bsigma^2)=-\frac{1}{2}(\hat\bz^\T\hat\bz+\frac{1}{\sigma^2}\parallel \bx-\bW\hat\bz \parallel^2)+\mbox{const}.\label{app:marg}
\end{equation}

\subsection{Analysis of $f_{\alpha}(\delta)$ \label{app:f_behavior}}

Let us  define
\begin{align}
f_{\alpha}(\delta)=\frac{1}{2}(\hat\bz^\T(\delta)\hat\bz(\delta)+\frac{1}{\alpha}\parallel \bx-\bW(\delta)\hat\bz(\delta) \parallel^2),\label{ap:eq:quad}
\end{align}
and study its properties. 

We observe that this function depends on $\bx$ but we do not make this dependency explicit in order to simplify the notation. Note also that we will end up studying the basic properties of the associated quadratic form.

\begin{theorem}
Let us consider $f_{\alpha}(\delta)$ defined in Eq. (\ref{ap:eq:quad}) with $\alpha<\lambda_P$ and  $\delta\in [0,\lambda_P)$. Then
\begin{itemize}
    \item $f_{\alpha}(\delta)$ is a convex function on $\delta$ with minimum at $\alpha/2$,
    \item and  $f_{\alpha}(0)=f_{\alpha}(\alpha)$.
\end{itemize}
\begin{proof}
Therefore, 
\begin{align}
&f_{\alpha}(\delta)\nonumber\\
\!&=\frac{1}{2}(\bx^\T\bU(\bL\!-\!\delta\bI)\bL^{\!-2}\bU^\T\bx\!+\!\frac{1}{\alpha}\parallel\bx\!-\!\bU(\bL\!-\!\delta\bI)\bL^{\!-1}\bU^\T\bx\parallel^2)\nonumber\\
\!&=\frac{1}{2}(\bx^\T\bU(\bL-\delta\bI)\bL^{-2}\bU^\T\bx\nonumber\\
\!&+\frac{1}{\alpha}(\bx^T\bx\!+\!\bx^T\bU(\bL\!-\!\delta\bI)^2\bL^{\!-2}\bU^\T\bx\nonumber\!-\!2\bx^\T\bU(\bL\!-\!\delta\bI)\bL^{\!-1}\bU^\T\bx))
\end{align}

and

\begin{align}
f^\prime_{\alpha}(\delta)&=\frac{1}{2}(-\bx^\T\bU\bL^{-2}\bU^\T\bx\\
&+\frac{1}{\alpha}(-2\bx^T\bU(\bL-\delta\bI)\bL^{-2}\bU^\T\bx+2\bx^\T\bU\bL^{-1}\bU^\T\bx)).\nonumber
\end{align}
Furthermore, we have the following identity for the sum of the matrices involved in $f^\prime_{\alpha}(\delta)$,
\begin{equation}
-\bL^{-2}+\frac{2}{\alpha}(-(\bL-\delta\bI)\bL^{-2}+\bL^{-1})=\bL^{-2}[-\bI+\frac{2}{\alpha}\delta \bI]
\end{equation}
from which we can see that the sum is the zero matrix  iff $\delta=\alpha/2$.

We also note that 
\begin{equation}
f_{\alpha}^{\prime\prime}(\delta)=\frac{1}{2\alpha}\bx^T\bU\bL^{-2}\bU^\T\bx\ge 0
\end{equation}

So, $f_\alpha(\delta)$ is a convex quadratic function on $\delta$ whose minimum is achieved at $\delta=\alpha/2$. 

Furthermore, using the Taylor expansion around the minimum it follows that
\begin{equation}
f_{\alpha}(\delta)=f_{\alpha}(\frac{\alpha}{2})+\frac{1}{2}(\delta-\frac{\alpha}{2})^2f^{\prime\prime}_{\alpha}(\frac{\alpha}{2})
\end{equation}
and so
$f_{\alpha}(0)=f_{\alpha}(\alpha)$.

In summary, $f_{\alpha}^{\prime\prime}(\delta)$ is convex, its minimum value is achieved at $\delta=\alpha/2$, and $f_{\alpha}(0)=f_{\alpha}(\alpha)$.
\end{proof}
\end{theorem}

\bibliographystyle{./bibliography/IEEEtran}
\bibliography{./bibliography/IEEEabrv,./bibliography/biblio}

\begin{thebibliography}{10}
\providecommand{\url}[1]{#1}
\csname url@samestyle\endcsname
\providecommand{\newblock}{\relax}
\providecommand{\bibinfo}[2]{#2}
\providecommand{\BIBentrySTDinterwordspacing}{\spaceskip=0pt\relax}
\providecommand{\BIBentryALTinterwordstretchfactor}{4}
\providecommand{\BIBentryALTinterwordspacing}{\spaceskip=\fontdimen2\font plus
\BIBentryALTinterwordstretchfactor\fontdimen3\font minus
  \fontdimen4\font\relax}
\providecommand{\BIBforeignlanguage}[2]{{%
\expandafter\ifx\csname l@#1\endcsname\relax
\typeout{** WARNING: IEEEtran.bst: No hyphenation pattern has been}%
\typeout{** loaded for the language `#1'. Using the pattern for}%
\typeout{** the default language instead.}%
\else
\language=\csname l@#1\endcsname
\fi
#2}}
\providecommand{\BIBdecl}{\relax}
\BIBdecl

\bibitem{Gartner2018:Online}
R.~Contu, D.~Kish, C.~Canales, S.~Deshpande, E.~Kim, and D.~Gardner, ``Forecast
  analysis: Information security, worldwide, 2q18 update,''
  url{https://www.gartner.com/en/documents/3889055}, 2018.

\bibitem{nytimes2021:online}
{New York Times}, ``{Blackout Hits Iran Nuclear Site in What Appears to Be
  Israeli Sabotage},''
  {https://www.nytimes.com/2021/04/11/world/middleeast/iran-nuclear-natanz.html},
  2021.

\bibitem{Nassif_survey}
A.~B. Nassif, M.~A. Talib, Q.~Nasir, and F.~M. Dakalbab, ``Machine learning for
  anomaly detection: A systematic review,'' \emph{IEEE Access}, vol.~9, pp.
  78\,658--78\,700, 2021.

\bibitem{kwon2019survey}
D.~Kwon, H.~Kim, J.~Kim, S.~C. Suh, I.~Kim, and K.~J. Kim, ``A survey of deep
  learning-based network anomaly detection,'' \emph{Cluster Computing},
  vol.~22, no.~1, pp. 949--961, 2019.

\bibitem{chalapathy2019deep}
R.~Chalapathy and S.~Chawla, ``Deep learning for anomaly detection: A survey,''
  2019.

\bibitem{ruff2021unifying}
L.~Ruff, J.~R. Kauffmann, R.~A. Vandermeulen, G.~Montavon, W.~Samek, M.~Kloft,
  T.~G. Dietterich, and K.-R. M{\"u}ller, ``A unifying review of deep and
  shallow anomaly detection,'' \emph{Proceedings of the IEEE}, 2021.

\bibitem{lakhina2004diagnosing}
A.~Lakhina, M.~Crovella, and C.~Diot, ``Diagnosing network-wide traffic
  anomalies,'' \emph{ACM SIGCOMM computer communication review}, vol.~34,
  no.~4, pp. 219--230, 2004.

\bibitem{lakhina2004characterization}
------, ``Characterization of network-wide anomalies in traffic flows,'' in
  \emph{Proceedings of the 4th ACM SIGCOMM conference on Internet measurement},
  2004, pp. 201--206.

\bibitem{macia2016hierarchical}
G.~Maci{\'a}-Fern{\'a}ndez, J.~Camacho, P.~Garc{\'\i}a-Teodoro, and R.~A.
  Rodr{\'\i}guez-G{\'o}mez, ``{Hierarchical PCA-based multivariate statistical
  network monitoring for anomaly detection},'' in \emph{2016 IEEE international
  workshop on information forensics and security (WIFS)}.\hskip 1em plus 0.5em
  minus 0.4em\relax IEEE, 2016, pp. 1--6.

\bibitem{Hotelling}
\BIBentryALTinterwordspacing
H.~Hotelling, ``Multivariate quality control,'' \emph{Techniques of Statistical
  Analysis}, 1947. [Online]. Available:
  \url{https://ci.nii.ac.jp/naid/10021322508/en/}
\BIBentrySTDinterwordspacing

\bibitem{Camacho2019}
J.~Camacho, J.~M. García-Jiménez, N.~M. Fuentes-García, and
  G.~Maciá-Fernández, ``Multivariate big data analysis for intrusion
  detection: 5 steps from the haystack to the needle,'' \emph{Computers \&
  Security}, vol.~87, p. 101603, 2019.

\bibitem{tipping1999probabilistic}
M.~E. Tipping and C.~M. Bishop, ``Probabilistic principal component analysis,''
  \emph{Journal of the Royal Statistical Society: Series B (Statistical
  Methodology)}, vol.~61, no.~3, pp. 611--622, 1999.

\bibitem{kingma2014}
D.~P. Kingma and M.~Welling, ``Auto-encoding variational bayes,'' 2014.

\bibitem{Xu2018VAE}
H.~Xu, W.~Chen, N.~Zhao, Z.~Li, J.~Bu, Z.~Li, Y.~Liu, Y.~Zhao, D.~Pei, Y.~Feng,
  J.~Chen, Z.~Wang, and H.~Qiao, ``{Unsupervised Anomaly Detection via
  Variational Auto-Encoder for seasonal KPIs in Web Applications},'' in
  \emph{Proc. of the 2018 World Wide Web Conference}, 2018, pp. 187--196.

\bibitem{camacho2017traffic}
J.~Camacho, P.~Garc{\'\i}a-Teodoro, and G.~Maci{\'a}-Fern{\'a}ndez, ``Traffic
  monitoring and diagnosis with multivariate statistical network monitoring: a
  case study,'' in \emph{2017 IEEE security and privacy workshops (SPW)}.\hskip
  1em plus 0.5em minus 0.4em\relax IEEE, 2017, pp. 241--246.

\bibitem{Soule2005PCA}
A.~Soule, A.~Lakhina, N.~Taft, K.~Papagiannaki, K.~Salamantian, M.~C.
  Antonio~Nucci, and C.~Diot, ``Traffic matrices: balancing measurements,
  inference and modeling,'' vol.~33, no.~1.\hskip 1em plus 0.5em minus
  0.4em\relax USENIX Association, Berkeley, CA, USA, 2005, pp. 331--344.

\bibitem{Zhang2005PCA}
Y.~Zhang, Z.~Ge, A.~Greenberg, and M.~Roughan, ``Network anomography,'' in
  \emph{Proceedings of the 5th ACM SIGCOMM conference on Internet Measurement
  (IMC'05)}.\hskip 1em plus 0.5em minus 0.4em\relax USENIX Association,
  Berkeley, CA, USA, 2005, pp. 317--330.

\bibitem{Huang2006PCA}
L.~Huang, X.~Nguyen, M.~Garofalakis, M.~Jordan, A.~Joseph, and N.~Taft,
  ``In-network pcas and anomaly detection,'' in \emph{Proceedings of Neural
  Information Processing Systems (NIPS)}.\hskip 1em plus 0.5em minus
  0.4em\relax NIPS Foundation, 2006.

\bibitem{Ringberg2007LimitationsPCA}
H.~Ringberg, A.~Soule, J.~Rexford, and C.~Diot, ``{Sensitivity of PCA for
  traffic anomaly detection},'' \emph{SIGMETRICS. Perform Eval. Rev.}, vol.~35,
  no.~1, pp. 109--120, 2007.

\bibitem{Rubistein2009RobustPCA}
B.~Rubinstein, B.~Nelson, L.~Huang, A.~Joseph, S.~hon Lau, R.~Satish, N.~Taft,
  , and J.~Tygar, ``Antidote: understanding and defending against poisoning of
  anomaly detectors,'' 2009.

\bibitem{Wang2012PCA}
Z.~Wang, K.~Hu, K.~Xu, Y.~Baolin, and X.~Dong, ``Structural analysis of network
  traffic matrix via relaxed principal component pursuit,'' \emph{Computer
  Networks}, vol.~56, pp. 2049--2067, 2012.

\bibitem{Pascoal2012RobustPCA}
C.~Pascoal, M.~{De Oliveira}, R.~Valadas, P.~Filzmoser, P.~Salvador, and
  A.~Pacheco, ``Robust feature selection and robust pca for internet traffic
  anomaly detection,'' 2012, pp. 1755--1763.

\bibitem{Brauckhoff2007LimitationsPCA}
D.~Brauckhoff, K.~Salamantian, and M.~May, ``{Applying PCA for Traffic Anomaly
  Detection: Problems and Solutions},'' in \emph{Proceedings of IEEE INFOCOM},
  2009, pp. 2886--2870.

\bibitem{K&MG96}
T.~Kourti and J.~F. MacGregor, ``{Multivariate SPC methods for process and
  product monitoring},'' \emph{Journal of Quality Technology}, vol.~28, no.~4,
  1996.

\bibitem{ferrer_latent:2014}
A.~Ferrer, ``Latent {Structures}-{Based} {Multivariate} {Statistical} {Process}
  {Control}: {A} {Paradigm} {Shift},'' \emph{Quality Engineering}, vol.~26, pp.
  72--91, Jan. 2014, publisher: Taylor \& Francis \_eprint:
  https://doi.org/10.1080/08982112.2013.846093.

\bibitem{G&F&W11}
J.~M. González-Martínez, A.~Ferrer, and J.~A. Westerhuis, ``{Real-time
  synchronization of batch trajectories for on-line multivariate statistical
  process control using Dynamic Time Warping},'' \emph{Chemometrics and
  Intelligent Laboratory Systems}, vol. 105, pp. 195--206, 2011.

\bibitem{Soufiane2021}
S.~Soufiane, R.~Magán-Carrión, I.~Medina-Bulo, and H.~Bouden, ``Preserving
  authentication and availability security services through multivariate
  statistical network monitoring,'' \emph{Journal of Information Security and
  Applications}, vol.~58, p. 102785, 2021.

\bibitem{LopezMartin2017CVAE}
M.~López-Martin, B.~Carro, A.~Sanchez-Esguevillas, and J.~Lloret,
  ``Conditional variational autoencoder for prediction and feature recovery
  applied to intrusion detection in iot,'' \emph{Sensors}, vol.~17, p. 1967,
  2017.

\bibitem{An2015VAE}
J.~An and S.~Cho, ``Variational autoencoder based anomaly detection using
  reconstruction probability,'' \emph{Technical Report}, pp. 1--18, 2015.

\bibitem{Zavrak2020VAE}
S.~Zavrak and M.~Iskefiyelo, ``Anomaly-based intrusion detection from network
  flow features,'' \emph{IEEE Access}, vol.~8, pp. 108\,346--108\,358, 2020.

\bibitem{Solch2016VAE}
M.~Sölch, J.~Bayer, M.~Ludersdorfer, and P.~van~der Smagt, ``Variational
  inference for on-line anomaly detection in high-dimensional time series,'' in
  \emph{International Conference on Machine Learning anomaly detection
  workshop}.\hskip 1em plus 0.5em minus 0.4em\relax IEEE, 2016.

\bibitem{bishop:2006}
C.~M. Bishop, \emph{Pattern Recognition and Machine Learning (Information
  Science and Statistics)}.\hskip 1em plus 0.5em minus 0.4em\relax Berlin,
  Heidelberg: Springer-Verlag, 2006.

\bibitem{MACIAFERNANDEZ_UGR16}
\BIBentryALTinterwordspacing
G.~Maciá-Fernández, J.~Camacho, R.~Magán-Carrión, P.~García-Teodoro, and
  R.~Therón, ``{UGR‘16: A new dataset for the evaluation of
  cyclostationarity-based network IDSs},'' \emph{Computers \& Security},
  vol.~73, pp. 411--424, 2018. [Online]. Available:
  \url{https://www.sciencedirect.com/science/article/pii/S0167404817302353}
\BIBentrySTDinterwordspacing

\bibitem{Ring_survey}
\BIBentryALTinterwordspacing
M.~Ring, S.~Wunderlich, D.~Scheuring, D.~Landes, and A.~Hotho, ``A survey of
  network-based intrusion detection data sets,'' \emph{Computers \& Security},
  vol.~86, pp. 147--167, 2019. [Online]. Available:
  \url{https://www.sciencedirect.com/science/article/pii/S016740481930118X}
\BIBentrySTDinterwordspacing

\bibitem{Castillo_DoSDatasets}
\BIBentryALTinterwordspacing
M.~Catillo, A.~Pecchia, M.~Rak, and U.~Villano, ``Demystifying the role of
  public intrusion datasets: A replication study of dos network traffic data,''
  \emph{Computers \& Security}, vol. 108, p. 102341, 2021. [Online]. Available:
  \url{https://www.sciencedirect.com/science/article/pii/S0167404821001656}
\BIBentrySTDinterwordspacing

\bibitem{GARCIA2014Botnet}
\BIBentryALTinterwordspacing
S.~García, M.~Grill, J.~Stiborek, and A.~Zunino, ``An empirical comparison of
  botnet detection methods,'' \emph{Computers \& Security}, vol.~45, pp.
  100--123, 2014. [Online]. Available:
  \url{https://www.sciencedirect.com/science/article/pii/S0167404814000923}
\BIBentrySTDinterwordspacing

\end{thebibliography}
\end{document}